\documentclass[letterpaper]{article} % DO NOT CHANGE THIS
\usepackage{aaai2027}  % DO NOT CHANGE THIS
\usepackage[hyphens]{url}  % DO NOT CHANGE THIS
\usepackage{graphicx} % DO NOT CHANGE THIS
\usepackage{natbib}  % DO NOT CHANGE THIS AND DO NOT ADD ANY OPTIONS TO IT
\usepackage{caption} % DO NOT CHANGE THIS AND DO NOT ADD ANY OPTIONS TO IT
\usepackage{algorithm}
\usepackage{algorithmic}

\usepackage{newfloat}
\usepackage{listings}
\DeclareCaptionStyle{ruled}{labelfont=normalfont,labelsep=colon,strut=off} % DO NOT CHANGE THIS
\floatstyle{ruled}
\newfloat{listing}{tb}{lst}{}
\floatname{listing}{Listing}

\usepackage{booktabs}

\usepackage{comment}
\usepackage{xcolor}
\usepackage[most]{tcolorbox}
\usepackage{newtxtext}
\usepackage{pifont}
\usepackage{enumitem}
\usepackage{booktabs}
\usepackage{multirow}
\usepackage{graphicx}
\usepackage{amsmath}
\usepackage{amssymb}

\usepackage{xspace}

\newcommand{\ours}{\textsf{DeMTS}\xspace}
\usepackage{booktabs}
\usepackage{tabularx}
\usepackage{array}
\usepackage{multirow}
\usepackage[table]{xcolor}
\usepackage{subcaption}

\usepackage[most]{tcolorbox}
\usepackage{listings}
\tcbuselibrary{listings,breakable}

\tcbset{
promptstyle/.style={
enhanced,
breakable,
listing only,
listing engine=listings,
width=\columnwidth,
boxrule=0.6pt,
arc=1mm,
left=1mm,
right=1mm,
top=1mm,
bottom=1mm,
before skip=6pt,
after skip=6pt,
fonttitle=\bfseries\small,
listing options={
basicstyle=\ttfamily\scriptsize,
numbers=none,
breaklines=true,
breakatwhitespace=false,
columns=fullflexible,
keepspaces=true,
showstringspaces=false,
upquote=true,
tabsize=2,
xleftmargin=0pt,
xrightmargin=0pt
}
}
}

\newtcblisting{generationprompt}[1]{
promptstyle,
colback=blue!5,
colframe=blue!50,
title={#1}
}

\newtcblisting{chatprompt}[1]{
promptstyle,
colback=green!5,
colframe=green!50!black,
title={#1}
}

\newtcblisting{judgeprompt}[1]{
promptstyle,
colback=orange!5,
colframe=orange!70!black,
title={#1}
}

\title{DeMTS: Denoising Trajectories as Multivariate Time Series for \\ Hallucination Detection in Diffusion Language Models}
\author{
    Xin Zhang\textsuperscript{\rm 1}\equalcontrib,
    Yili Wang\textsuperscript{\rm 1}\equalcontrib,
    Yue Tan\textsuperscript{\rm 2},
    Xin He\textsuperscript{\rm 1},\\
    Yanyu Qian\textsuperscript{\rm 3},
    Yixin Liu\textsuperscript{\rm 2}\corresponding,
    Yi Chang\textsuperscript{\rm 1},
    Shirui Pan\textsuperscript{\rm 2},
    Xin Wang\textsuperscript{\rm 1}\corresponding
}
\affiliations{
    \textsuperscript{\rm 1}Jilin University, China 
    \textsuperscript{\rm 2}Griffith University, Australia  
    \textsuperscript{\rm 3}Nanyang Technological University, Singapore

}

\begin{document}
\maketitle

\begin{abstract}
Diffusion large language models (D-LLMs) have emerged as a promising paradigm for text generation. 
However, similar to autoregressive LLMs, D-LLMs remain vulnerable to hallucinations, where fluent outputs may contain factually incorrect or unsupported content. 
Although existing hallucination detection methods for D-LLMs attempt to leverage uncertainty trajectories of the denoising process to better identify hallucination signals, they typically compress the trajectories along either the temporal or token dimension, overlooking the useful information encoded in the complete two-dimensional token-step structure. Consequently, they may fail to capture hallucination-relevant patterns, such as inconsistent convergence and cross-token fault propagation, leading to suboptimal detection performance. 
To bridge this gap, we propose a D-LLM hallucination detection framework that formulates the \textbf{De}noising trajectories as \textbf{M}ultivariate \textbf{T}ime \textbf{S}eries over learnable latent variables (\ours for short). 
\ours employs a trajectory-preserving token-to-variable assignment module to convert token signals into stable latent variables. 
Based on these variables, we propose dynamic multivariate temporal modeling to progressively integrate inter-variable dependency modeling with temporal encoding for hallucination prediction. 
Extensive experiments on two D-LLMs backbones and three benchmarks demonstrate that \ours outperforms existing hallucination detection methods while maintaining strong robustness, efficiency, and cross-task transferability. 
\end{abstract}


\section{Introduction}
Diffusion large language models (D-LLMs) have recently gained increasing attention as an emerging paradigm for efficient and high-quality text generation~\cite{llada,llada100b,dream}. Rather than decoding tokens strictly from left to right~\cite{mehri2018middle,gu2019insertion,ghazvininejad2019mask}, D-LLMs generate a sequence through iterative denoising. Despite their advantages over autoregressive LLMs, D-LLMs remain vulnerable to hallucinations, where fluent outputs may contain factually incorrect or unsupported content~\cite{hallucination,hallucination2}.
Detecting such errors from output-level evidence (i.e., predictive uncertainty of output tokens) is often insufficient~\cite{shoby2026overthinking,badave2026beyond,bhatnagar2026drift}, because hallucination-related cues may appear during the denoising process in the form of uncertainty rebound or inconsistent refinement among token~\cite{hallucinationofDLLM1,hallucinationofDLLM2}. 
This motivates D-LLM-specific hallucination detection methods that explicitly model denoising trajectories, rather than solely relying on static features of the generated output~\cite{chen2026hallusae,zhang2025icr}.

Recent studies have begun to exploit predictive uncertainty throughout the denoising trajectory, typically quantified by token-level entropy, for hallucination detection in D-LLMs~\cite{hallucinationofDLLM3,mtssurvey}. 
One line of work focuses on \textbf{step-level} evidence, which compresses the denoising trajectory along the diffusion-step dimension by selecting or reweighting the steps whose uncertainty patterns are most informative for hallucination prediction~\cite{tracedet,weng2026tre}. 
Another line models \textbf{token-level} uncertainty dynamics, which compresses the trajectory along the token dimension by identifying hallucination-indicative tokens and using their denoising dynamics as indicators of factual reliability~\cite{dynhd}.
Although these methods demonstrate the usefulness of uncertainty at intermediate denoising states, both lines of works compress the original trajectory along one dimension before detection.  
As a result, the token-step structure of D-LLM denoising trajectories, which jointly captures how token-level uncertainty evolves across diffusion steps, is not fully preserved, potentially leading to information loss for hallucination detection.

\begin{figure}[t]
\centering
\includegraphics[width=\linewidth]{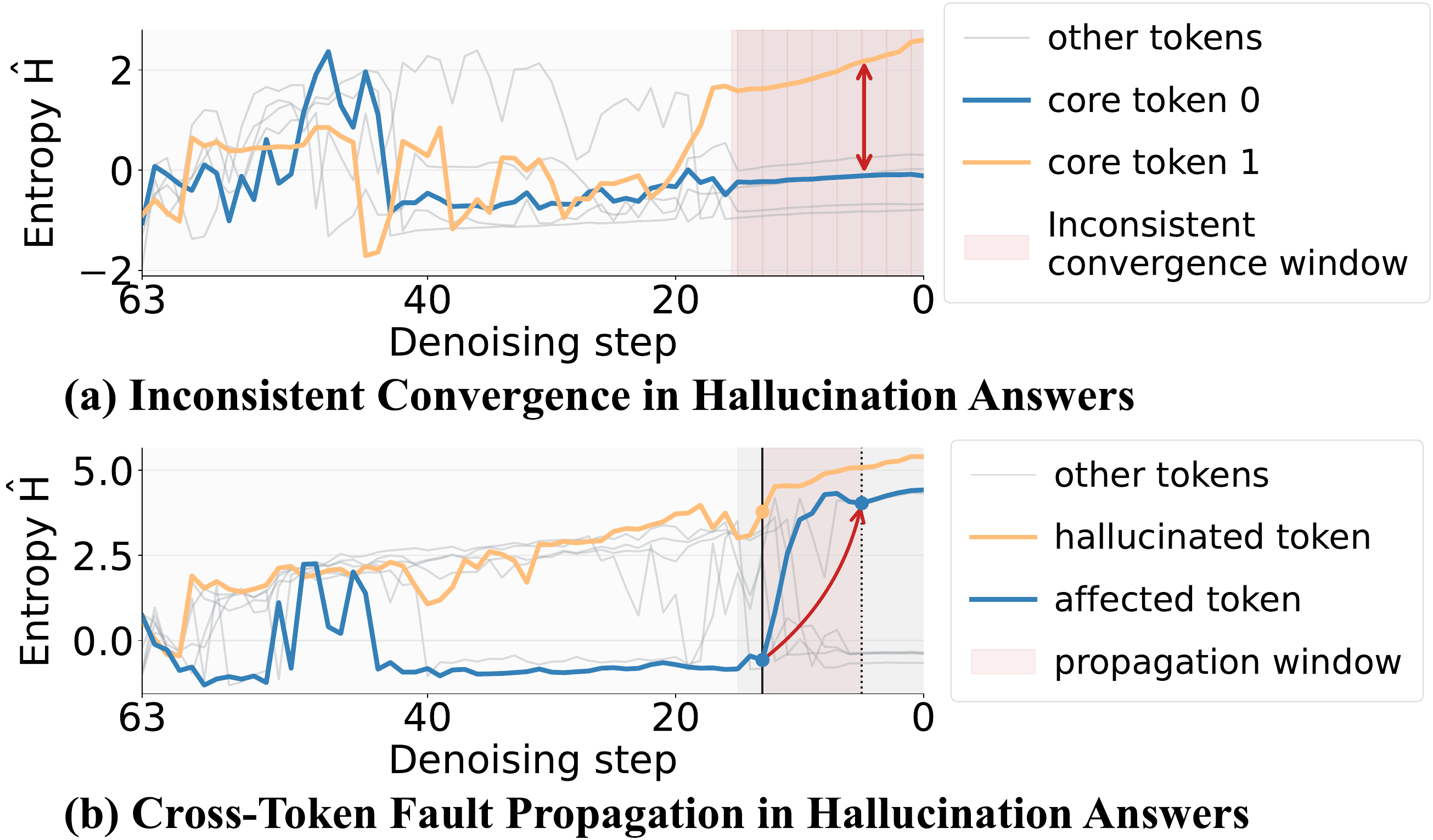}
\vspace{-6mm}
\caption{Observed patterns in denoising trajectories.}
\label{figure:phenomenon}
\vspace{-3mm}
\end{figure}

To investigate whether compressing denoising trajectories may overlook critical cues for hallucination detection, we examine the evolution of token-level uncertainty under different Q\&A scenarios, and then identify two representative patterns that require explicit modeling of token-step uncertainty evidence. 
\textbf{Pattern~\ding{182}:~Inconsistent Convergence in Hallucination Answers} (Fig.~\ref{figure:phenomenon}a). 
In hallucinated responses, token uncertainties may converge to markedly different levels during the late denoising stage: some tokens become highly confident, while others remain relatively uncertain. Such inconsistent convergence across tokens reveals abnormal inter-token relations and provides hallucination evidence beyond any individual token.
\textbf{Pattern~\ding{183}:~Cross-Token Fault Propagation in Hallucination Answers} (Fig.~\ref{figure:phenomenon}b). During the denoising process of some hallucination answers, an unstable or incorrect token state can affect the decoding of other contextually connected tokens, causing their entropy trajectories to deviate together from a coherent factual path. 
Capturing this propagative pattern for hallucination detection requires explicit modeling of dynamic dependencies among token trajectories, further highlighting the necessity of preserving the full token-step uncertainty structure. Recognizing the importance of two-dimensional evidence modeling, we pose a central research question: 

\begin{tcolorbox}[
colback=blue!3,
colframe=blue!45!black,
boxrule=0.8pt,
arc=4pt,
left=2pt,
right=2pt,
top=2pt,
bottom=2pt
]
{How can we explicitly model \textit{\textbf{token-step uncertainty trajectories}} to capture cross-token dependencies and temporal dynamics for D-LLM hallucination detection?}
\end{tcolorbox}

A promising solution to this question is to formulate the token-step uncertainty trajectory as a \textbf{multivariate time series}, which naturally preserves its two-dimensional structure: 
denoising steps define the temporal axis, while token-associated uncertainty signals serve as multiple interacting variables. %This perspective enables hallucination detection to move beyond isolated denoising stages or individual token dynamics and instead capture how uncertainty signals jointly evolve throughout generation. 
Although this formulation is natural, directly applying existing approaches for multivariate time series to D-LLM hallucination detection is non-trivial, since they typically assume that each variable has a stable identity over time and among different variables~\cite{timefilter,crossformer}. This assumption does not hold for D-LLMs, where the semantic role of a potential variable (i.e. token position) may vary across samples and change during denoising process, thus giving rise to two key challenges. 
\textbf{\emph{Challenge~1:~Stable Variable Construction.}} Raw token positions cannot directly serve as variables, as they lack consistent semantic identities. They must be reorganized into semantically stable variables while preserving fine-grained denoising dynamics relevant to hallucination detection.
\textbf{\emph{Challenge~2:~Relational Temporal Modeling.}} The temporal dynamics of the constructed variables are coupled with evolving dependencies among them. In this case, the hallucination detector must jointly capture both aspects, as hallucination cues can manifest as inconsistent convergence or propagate across related trajectories.

To address these challenges, we propose \ours, a hallucination detection framework that formulates D-LLM \textbf{De}noising trajectories as \textbf{M}ultivariate \textbf{T}ime \textbf{S}eries over learnable latent variables. To handle \textbf{\emph{Challenge~1}}, \ours introduces a \textbf{trajectory-preserving Token-to-Variable assignment} (T2V) module, which reorganizes token uncertainty trajectories into latent variables with more consistent identities, considering uncertainty states, contextual semantics, and positional structure. A trajectory-preservation constraint is further imposed to retain local denoising variations during token-to-variable assignment. To address \textbf{\emph{Challenge~2}}, \ours introduces \textbf{Dynamic Multivariate Temporal Modeling} (DMTM), which jointly captures relational and temporal patterns over the constructed latent variables. It performs step-adaptive inter-variable interaction to suppress unreliable early-step dependencies and strengthen interactions among stabilized variables, followed by variable-wise temporal modeling to encode the evolution of each latent variable. In this way, \ours captures both hallucination-related temporal dynamics and cross-token dependencies. In summary, our contributions are as follows:

\begin{itemize}
\item We formulate D-LLMs hallucination detection as a multivariate time series learning problem, and identify two key challenges in adapting existing time series approaches to denoising trajectories. 

\item We propose \ours, a trajectory-based detector which transforms semantically unstable token signals into stable latent variables and jointly captures dynamic inter-variable dependencies and variable-wise temporal evolution.

\item Experiments on two D-LLM backbones and three benchmarks show that \ours consistently outperforms baselines, while remaining robust, transferable, and efficient.

\end{itemize}

\section{Related Works}
\noindent\textbf{Hallucination Detection in D-LLMs.} 
The iterative denoising process of D-LLMs~\cite{diffusionmodel} provides rich trajectory-level uncertainty evidence for hallucination detection~\cite{hallucinationofDLLM3}. 
One line of existing methods focuses on \textit{step-oriented evidence selection}~\cite{tracedet,tdgnet,hive}, which identifies or reweights informative denoising stages and uses the selected trajectory segments for hallucination prediction. The second line focuses on \textit{dynamics-oriented trajectory modeling}~\cite{dynhd}, which detects hallucinations by modeling abnormal token evolution during denoising. They show that hallucination evidence can emerge from discriminative denoising steps or dynamic trajectory patterns. 
Differently, \ours models D-LLMs two-dimensional denoising trajectories as multivariate time series for hallucination detection.

\noindent\textbf{Multivariate Time Series Learning.} 
Deep learning for multivariate time series has been widely studied to capture evolving patterns across multiple variables~\cite{mts1,mtssurvey3,mtssurvey2}. Existing methods are commonly categorized by their channel modeling strategies. Channel-independent methods model each variable separately to preserve variable-specific temporal patterns~\cite{crossformer,mtscd1,shen2026raising}, while channel-dependent methods jointly model all variables to capture global inter-variable correlations~\cite{mtsci1,zhao2026fedcigar,mtsci2}. More recent channel-partial methods seek a balance between these two views by allowing each variable to interact only with relevant variables~\cite{mtscp1,mtscp2}. These strategies assume that variables have stable identities across samples and time~\cite{vcformer,tqnet}. However, in D-LLM denoising trajectories, token positions may correspond to changing semantic roles during generation. Therefore, \ours converts changing token signals into stable latent variables for temporal and inter-variable modeling. Detailed literature review please refer to Appendix~\ref{app:rw}.

\section{Preliminary}
\noindent\textbf{Diffusion Large Language Models.}
Given an input query $\mathbf{q}$, the diffusion large language models (D-LLMs) generates a fixed-length response through iterative denoising. Let
$\mathbf{r}^{(t)}=(r_{1}^{(t)},r_{2}^{(t)},\ldots,r_{N}^{(t)})\in\mathcal{V}^{N}$
denote the intermediate response at denoising step
$t\in\{T,\ldots,0\}$, where $N$ is the sequence length and $\mathcal{V}$ is the vocabulary. Starting from a highly masked sequence $\mathbf{r}^{(T)}$, the model progressively reconstructs the response according to
$
\mathbf{r}^{(t-1)}\sim p_{\theta}\left(\mathbf{r}^{(t-1)}\mid\mathbf{r}^{(t)},\mathbf{q}\right),
$
until obtaining the final response $\mathbf{r}^{(0)}$.

At each denoising step $t$, the D-LLMs predicts a categorical distribution over the vocabulary for each token position $i$:
$\boldsymbol{\pi}_{i,t}=p_{\theta}\left(r_{i}^{(0)}\mid\mathbf{r}^{(t)},\mathbf{q}\right)\in\mathbb{R}^{|\mathcal{V}|}.$
The corresponding predictive uncertainty is measured by token entropy:
$H_{i,t}=-\sum_{v\in\mathcal{V}}\pi_{i,t}(v)\log\pi_{i,t}(v).$
Collecting token uncertainties across all denoising steps and positions yields the uncertainty trajectory
$\mathbf{H}=[H_{i,t}]\in\mathbb{R}^{N\times(T+1)},$
where the temporal axis corresponds to denoising steps and the token axis corresponds to sequence positions. We further denote by
$\mathbf{e}_{i,t}\in\mathbb{R}^{d_h}$
the contextual embedding of token position $i$ at step $t$, and collect them as
$\mathbf{E}\in\mathbb{R}^{N\times (T+1)\times d_h}$.
The uncertainty trajectory $\mathbf{H}$ provides the primary detection signal, while $\mathbf{E}$ supplies contextual semantics for constructing the latent-variable trajectory.

\noindent\textbf{Hallucination Detection in D-LLMs.}
Given the uncertainty trajectory $\mathbf{H}$, contextual trajectory $\mathbf{E}$, and the generated response, hallucination detection aims to determine whether the response contains factually incorrect or unsupported content. Let
$\mathcal{D}=\left\{(\mathbf{H}_{n},\mathbf{E}_{n},y_{n})\right\}_{n=1}^{M}$
denote the training set, where $M$ is the number of samples and
$y_n\in\{0,1\}$ is the hallucination label, with $y_n=1$ indicating a hallucinated response and $y_n=0$ indicating a factual response.
The objective is to learn a detector $f_{\phi}$ that maps the denoising trajectories to a hallucination probability: $\hat{y}_{n}=f_{\phi}\left(\mathbf{H}_{n},\mathbf{E}_{n}\right).$
The task is formulated as trajectory-based binary classification:
\begin{equation}
\min_{\phi}\frac{1}{M}\sum_{n=1}^{M}\mathcal{L}_{\mathrm{cls}}\left(y_{n},f_{\phi}(\mathbf{H}_{n},\mathbf{E}_{n})\right),
\end{equation}
where $\mathcal{L}_{\mathrm{cls}}$ denotes the binary cross-entropy loss.

\section{Methodology}

\begin{figure*}
\vspace{-4mm}
\centering
\includegraphics[width=\textwidth]{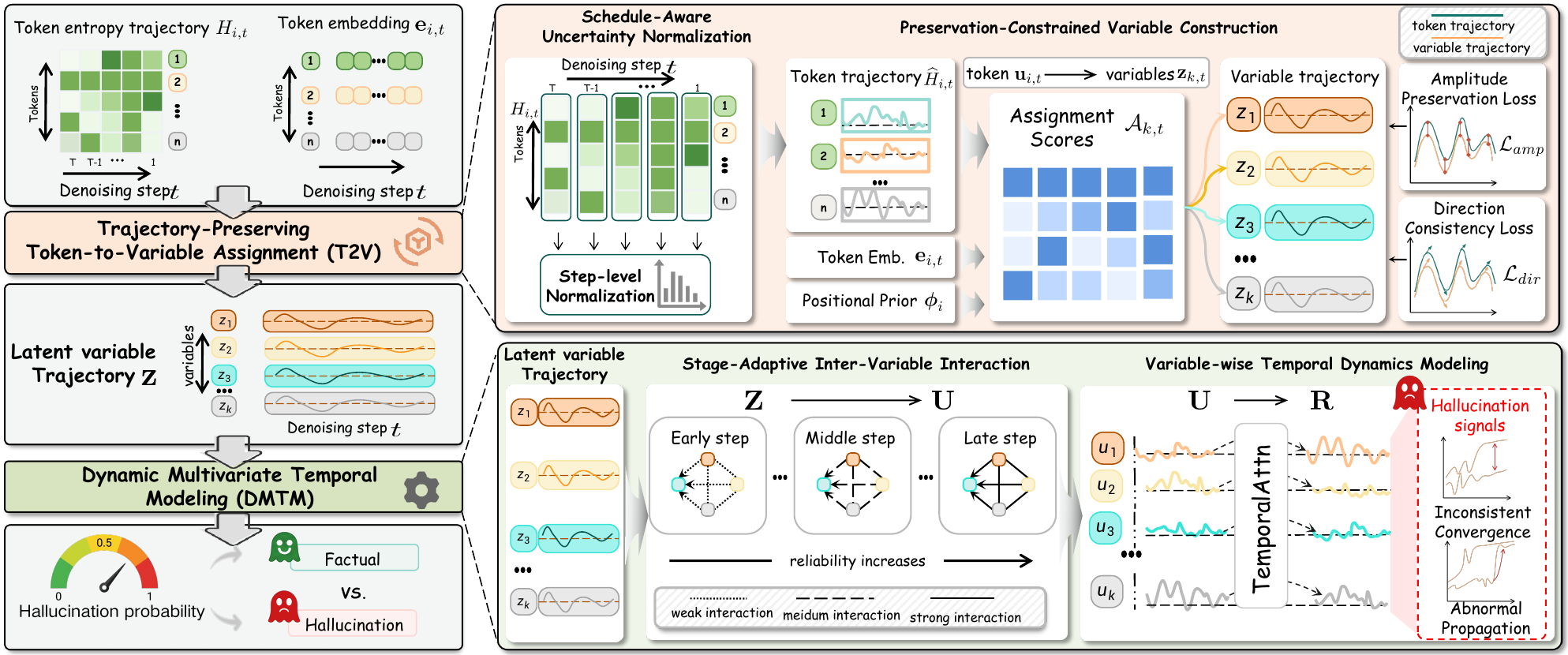}
\caption{Overview of \ours, which transforms token-level denoising signals into latent variables and jointly models their temporal evolution and inter-variable interactions for hallucination detection. }
\label{figure:framework}
\vspace{-1mm}
\end{figure*}

In this section, we provide an overview of \ours, a hallucination detection framework that models D-LLMs denoising trajectories as multivariate time series. \ours takes a token-step uncertainty trajectory as input which jointly preserves temporal evolution and cross-token relationships.
It first performs \textit{Trajectory-Preserving Token-to-Variable (T2V) Assignment} to map schedule-biased and semantically unstable token signals into latent variables with more consistent identities. 
Based on the resulting latent variable trajectories, \ours then applies \textit{Dynamic Multivariate Temporal Modeling} to jointly capture variable-wise temporal evolution and dynamically evolving interactions among latent variables throughout the denoising process. 
The learned trajectory representation is finally used for hallucination detection. The overall framework of \ours is illustrated in Fig.~\ref{figure:framework}.

\subsection{Trajectory-Preserving T2V Assignment}

The raw denoising trajectory of a D-LLM is organized by token positions, where each trajectory corresponds to the uncertainty evolution of a fixed sequence position. However, the same position may correspond to different semantic roles and evolve into different token contents during denoising, making it unsuitable as a variable for multivariate temporal modeling. 
To construct a more reliable variable axis, we propose \textit{Trajectory-Preserving Token-to-Variable Assignment} (T2V), which converts the position-indexed token trajectory into a latent variable-indexed trajectory, providing a more stable basis for multivariate time series learning. 

The goal of T2V is to learn how each token-level signal should contribute to $K$ latent variables with stable semantics while preserving its denoising dynamics. To this end, a {schedule-aware uncertainty normalization} scheme is introduced to reduce the global uncertainty bias induced by the denoising process. Then, a {latent variable axis} is constructed by assigning token-level signals to latent variables according to their uncertainty, semantic content, and positional structure. Moreover, T2V imposes a trajectory-preservation constraint on the assignment, preventing local hallucination cues from being smoothed out during token-to-variable aggregation.

\noindent\textbf{Schedule-Aware Uncertainty Normalization.}
Raw token uncertainty can be affected by the denoising schedule, where early steps are generally more uncertain and later steps become more stable. To reduce this step-wise bias, we normalize token uncertainty within each denoising step by
\begin{align} 
\small
\begin{gathered} 
\mu_t=\frac{1}{N}\sum_{i=1}^{N}H_{i,t}, \qquad \sigma_t=\sqrt{\frac{1}{N}\sum_{i=1}^{N}\left(H_{i,t}-\mu_t\right)^2},\\ 
\widehat{H}_{i,t}=\frac{H_{i,t}-\mu_t}{\sigma_t+\epsilon}, 
\end{gathered} 
\label{eq:normalization}
\end{align}
where $H_{i,t}$ denotes the raw uncertainty of token position $i$ at denoising step $t$, $\widehat{H}_{i,t}$ denotes the normalized uncertainty, and $\epsilon$ is a small constant. After normalization, $\widehat{H}_{i,t}$ can capture the relative uncertainty level of token $i$ within the uncertainty distribution at each denoising step, providing a schedule-calibrated signal for token-to-variable assignment.

\noindent\textbf{Variable Axis Construction.}
After normalization, the denoising trajectory remains indexed by raw token positions, preventing their direct use as stable variables for multivariate temporal modeling. Since a token position lacks a stable semantic identity in most scenarios, we construct a latent variable axis by learning a soft assignment from each token-level signal to $K$ latent variables, grouping semantically related token signals into latent variables with more consistent identities. 
% thereby reorganizing the position-indexed trajectory into a more stable variable-indexed trajectory.
In this way, the position-indexed trajectory is reorganized into a stable variable-indexed trajectory, providing a more reliable basis for subsequent procedures.

To construct a latent-variable axis with stable semantics, the assignment should depend on both uncertainty dynamics and token semantics. We therefore combine the normalized uncertainty with the contextual token embedding extracted from the D-LLMs via: 
\begin{align}
\small
\mathbf{u}_{i,t}
=
\mathrm{Linear}_{h}(\widehat{H}_{i,t})
+
\gamma\,\mathrm{Linear}_{e}(\mathbf{e}_{i,t}),
\label{eq:he}
\end{align}
where $\mathbf{u}_{i,t}\in  \mathbb{R}^{N \times(T+1) \times d_u}$ denotes the joint representation of token (i) at denoising step (t), encoding both its uncertainty state and contextual semantics for subsequent latent-variable assignment. Specifically, \(\mathbf{e}_{i,t}\) denotes the contextual embedding of token \(i\) at denoising step \(t\), \(\mathrm{Linear}_{h}(\cdot)\) and \(\mathrm{Linear}_{e}(\cdot)\) project the uncertainty and semantic signals into the same space, respectively, and \(\gamma\) controls the contribution of contextual information. 

Based on \(\mathbf{u}_{i,t}\), we further compute the content-conditioned assignment logits over \(K\) latent variables as: 
\begin{align}
\small
\mathbf{s}^{\mathrm{con}}_{i,t}
=
\mathcal{F}_{\mathrm{con}}(\mathbf{u}_{i,t}),
\label{eq:content_assignment_logits}
\end{align}
where $\mathcal{F}_{\mathrm{con}}(\cdot)$ is a learnable mapping function, and $\mathbf{s}^{\mathrm{con}}_{i,t}$ represents the assignment preference of token signal $(i,t)$ over the $K$ latent variables.

Although token positions cannot directly serve as stable variables due to their unstable semantic identities, positional information still provides useful structural cues for latent variable construction. 
Specifically, relative positions preserve the sequential order and local neighborhood relationships among tokens, offering structural guidance for grouping semantically related token-level signals into latent variables. 
% Relative positions encode the order and neighborhood relations among tokens, guiding the assignment module to group related token-level signals into latent variables. 
To capture structural cues at different spatial scales, we adopt a multi-frequency Fourier representation to encode each position: 
\begin{equation}
\small
\boldsymbol{\phi}_{i}
=
\left[
\sin\left(\frac{2\pi f i}{N}\right),
\cos\left(\frac{2\pi f i}{N}\right)
\right]_{f=1}^{F},
\label{eq:position}
\end{equation}
where \(F\) denotes the number of Fourier frequencies. 
Low-frequency components capture coarse global ordering, whereas high-frequency components preserve fine-grained local neighborhood relations, providing complementary structural information for latent-variable construction. 
We then map the Fourier representation into structural assignment logits: 
\begin{equation}
\small
\mathbf{s}^{\mathrm{str}}_{i}
=
\mathcal{F}_{\mathrm{str}}(\boldsymbol{\phi}_{i}),
\label{eq:structural_assignment_logits}
\end{equation}
where \(\mathcal{F}_{\mathrm{str}}(\cdot)\) is a learnable mapping function and \(\mathbf{s}^{\mathrm{str}}_{i}\) represents the structural assignment preference of position \(i\) over the \(K\) latent variables. 
Since positional structure should be adaptively injected according to the current token state, we introduce a content-adaptive gate
\begin{equation}
\small
\mathbf{g}_{i,t}
=
\sigma\left(
\mathcal{F}_{\mathrm{gate}}(\mathbf{u}_{i,t})
\right)
\in\mathbb{R}^{K},
\label{eq:content_adaptive_gate}
\end{equation}
to regulate their contribution. 
The final token-to-variable assignment weights are obtained by combining content-conditioned and gated structural assignment logits:
\begin{align}
\small
\mathbf{a}_{i,t}
=
\operatorname{Softmax}\left(
\mathbf{s}^{\mathrm{con}}_{i,t}
+
\mathbf{g}_{i,t}\odot\mathbf{s}^{\mathrm{str}}_{i}
\right)
\in\mathbb{R}^{K},
\label{eq:assignment}
\end{align}
where \(\mathbf{a}_{i,t}\) denotes the soft assignment distribution of token signal \((i,t)\) over the \(K\) latent variables. In this way, token semantics and positional structure are jointly considered for token-to-variable assignment. 

\noindent\textbf{Preservation-Constrained Variable Construction.}
Given the token-to-variable assignment, we further construct the variable-indexed trajectory through assignment-weighted aggregation, transforming token-level representations into latent-variable representations. We first define the aggregation operator for latent variable \(k\) at denoising step \(t\):
\begin{equation}
\small
\mathcal{A}_{k,t}(\mathbf{x}_{t})
=
\frac{
\sum_{i=1}^{N}
a_{i,t,k}\mathbf{x}_{i,t}
}{
\sum_{i=1}^{N}
a_{i,t,k}
+
\epsilon
},
\label{eq:aggregation_operator}
\end{equation}
where \(\mathbf{x}_{t}=\{\mathbf{x}_{i,t}\}_{i=1}^{N}\) denotes a generic token-indexed input to the aggregation operator instantiated as the joint token representation \(\mathbf{u}_{t}\). Moreover, \(\mathbf{a}_{i,t}=[a_{i,t,1},a_{i,t,2},\ldots,a_{i,t,K}]\in\mathbb{R}^{K}\) is the soft assignment distribution of token \(i\) over all latent variables at denoising step \(t\), and \(a_{i,t,k}\) denotes its \(k\)-th element. It measures the contribution of token-level input \(\mathbf{x}_{i,t}\) to the \(k\)-th latent variable, while \(\epsilon\) is a small constant for numerical stability. Using this operator, the latent variable \(k\) is written as
$
\mathbf{z}_{k,t}
=
\mathcal{A}_{k,t}(\mathbf{u}_{t})
\in\mathbb{R}^{d_u},
$
and the complete variable-indexed trajectory is
$
\mathbf{Z}
=
[\mathbf{z}_{k,t}]_{\substack{k=1,\ldots,K\\t=0,\ldots,T}}. 
$

A potential concern with assignment-weighted aggregation is that it may smooth out local denoising dynamics that are informative for hallucination detection. We therefore introduce a trajectory-preservation constraint to retain these dynamics. Specifically, we project the uncertainty change of token \(i\) into the representation space:
$
\Delta\widehat{\mathbf H}_{i,t} = \mathrm{Linear}_{h} (\widehat H_{i,t}-\widehat H_{i,t-1}).
$
Collecting all token-level variations gives
$
\Delta\widehat{\mathbf H}_{t}
=
[
\Delta\widehat{\mathbf H}_{1,t},
\ldots,
\Delta\widehat{\mathbf H}_{N,t}
]^{\top}.
$
We then aggregate the token-level variations into the latent-variable space:
$\Delta\mathbf Z_{k,t} = \mathcal A_{k,t} ( \Delta\widehat{\mathbf H}_{t}). $
The latent-variable variations are subsequently projected back to the token level:
$
\Delta\widetilde{\mathbf H}_{i,t}
=
\sum_{k=1}^{K}
a_{i,t,k}\Delta\mathbf Z_{k,t}.
$
This aggregation and reconstruction process constrains the T2V assignment to preserve token-level denoising dynamics during variable construction.

To preserve both the magnitude and temporal direction of these dynamics, we decompose the preservation constraint into two complementary terms. The first term is the \textbf{\textit{Amplitude Preservation Loss}}, which addresses amplitude smoothing, where token-to-variable aggregation weakens the magnitude of local uncertainty changes:
\begin{equation}
\small
\mathcal{L}_{amp}
=
\mathbb{E}_{i,t}
\left[
\left(
|\Delta\mathbf{\widehat{H}}_{i,t}|
-
|\Delta\mathbf{\widetilde{H}}_{i,t}|
\right)^2
\right].
\label{eq:amplitude_preservation}
\end{equation}
By matching the magnitude of the reconstructed variation to that of the original token-level variation, \(\mathcal{L}_{amp}\) encourages variable construction to retain the strength of local denoising dynamics. The second term is the \textbf{\textit{Direction Consistency Loss}}, which addresses direction distortion, where the reconstructed variation may evolve in the opposite temporal direction:
\begin{equation}
\small
\mathcal{L}_{dir}
=
\mathbb{E}_{i,t}
\left[
\max
\left(
0,
-
\Delta\mathbf{\widehat{H}}_{i,t}
\Delta\mathbf{\widetilde{H}}_{i,t}
\right)
\right].
\label{eq:direction_preservation}
\end{equation}
This penalty is activated when the original and reconstructed variations have opposite signs, thereby discouraging variable construction from reversing uncertainty trends. The final preservation constraint (with balance hyperparameters $\lambda$ ) is defined as
\begin{equation}
\small
\mathcal{L}_{pres}
=
\lambda_{amp}\mathcal{L}_{amp}
+
\lambda_{dir}\mathcal{L}_{dir}.
\label{eq:preservation_constraint}
\end{equation}

\subsection{Dynamic Multivariate Temporal Modeling}
With the latent-variable trajectory constructed by T2V, the remaining challenge is to model the temporal evolution with the evolving inter-variable dependency for hallucination detection. 
To this end, we propose \textit{Dynamic Multivariate Temporal Modeling} (DMTM), which progressively integrates inter-variable dependency modeling with variable-wise temporal encoding for hallucination prediction.

\noindent\textbf{Stage-Adaptive Inter-Variable Interaction.}
In D-LLM denoising trajectories, the dependencies among latent variables evolve throughout denoising and provide important relational evidence for hallucination detection. To model the dependencies, we aggregate cross-variable relational evidence through attention-based interaction. Specifically, we construct contextualized variable representation via:
\begin{equation}
\small
\mathbf{C}
=
\operatorname{Softmax}\left(
\frac{\mathbf{Q}\mathbf{K}^{\top}}{\sqrt{d}}
\right)\mathbf{V},
\label{eq:inter_variable_interaction}
\end{equation}
where $\mathbf{Q}=\mathbf{Z}\mathbf{W}_{Q}$, $\mathbf{K}=\mathbf{Z}\mathbf{W}_{K}$, and $\mathbf{V}=\mathbf{Z}\mathbf{W}_{V}$ denote the query, key, and value projections, respectively. The row-wise softmax normalizes the dependency scores for each latent variable across all variables, producing relation weights that specify how much information it should receive from the others at step $t$. Using these weights, each latent variable aggregates information from its related variables, adaptively capturing step-specific cross-variable dependencies during denoising. 

While the above interaction captures step-specific dependencies, their reliability varies with the noise level at each denoising step. 
In the D-LLMs denoising process, larger $t$ corresponds to earlier and noisier stages, where variable are less reliable and the estimated dependencies are more susceptible to spurious correlations. As $t$ decreases, the variables gradually stabilize, making their relational patterns increasingly representative of the underlying semantic dependencies~\cite{weng2026tre}. To calibrate cross-variable information fusion according to this reliability transition, we introduce a trajectory-reliability gate:
\vspace{-1mm}
\begin{equation}
\small
\mathbf{U}
=
\mathbf{Z}
+
\rho\mathbf{C},
\qquad
\rho=\frac{T-t}{T},
\label{eq:featuremerge}
\end{equation}
\vspace{-1mm}
where $\rho_t$ increases as the trajectory approaches the final step, assigning limited fusion strength to noisy early-step relations while progressively strengthening information exchange among stabilized variables. 
By limiting interaction at noisy early stages and strengthening it as the variables stabilize, this design reduces the influence of spurious dependencies and emphasizes more reliable semantic relations.

\newcolumntype{C}{>{\centering\arraybackslash}X}
\begin{table*}[ht!]
\vspace{-3mm}
\centering
\small
\setlength{\tabcolsep}{3pt}
\renewcommand{\arraystretch}{1.08}
\resizebox{1.9\columnwidth}{!}{
\begin{tabularx}{\textwidth}{c l l *{6}{C} c}
\toprule
\multirow{2}{*}{\textbf{Model}}
& \multicolumn{2}{c}{\multirow{2}{*}{\textbf{Method}}}
& \multicolumn{2}{c}{\textbf{TriviaQA}}
& \multicolumn{2}{c}{\textbf{HotpotQA}}
& \multicolumn{2}{c}{\textbf{CSQA}}
& \multirow{2}{*}{\textbf{Avg.}} \\

\cmidrule(lr){4-5}
\cmidrule(lr){6-7}
\cmidrule(lr){8-9}

& \multicolumn{2}{c}{}
& 64 & 128
& 64 & 128
& 64 & 128
& \\
\midrule

% ==================== LLaDA ====================
\multirow{10}{*}{
    \rotatebox[origin=c]{90}{\textbf{LLaDA-8B-Instruct}}
}
& \multirow{4}{*}{\shortstack{Output-based\\Methods}}
& Perplexity
& 47.6 & 50.4 & 51.2 & 49.3 & 65.0 & 65.6 & 54.9 \\

&
& LN-Entropy
& 53.5 & 54.6 & 54.7 & 54.8 & 64.4 & 64.6 & 57.8 \\

&
& Semantic Entropy
& 67.3 & 68.9 & 53.8 & 57.6 & 43.9 & 44.1 & 55.9 \\

&
& Lexical Similarity
& 59.0 & 62.5 & 57.1 & 64.2 & 60.7 & 57.3 & 60.1 \\

\cmidrule(lr){2-10}

& \multirow{3}{*}{\shortstack{Latent-based\\Methods}}
& EigenScore
& 66.9 & 69.2 & 59.2 & 64.7 & 60.6 & 58.5 & 63.2 \\

&
& CCS
& 54.2 & 57.1 & 55.8 & 57.6 & 58.5 & 50.5 & 55.6 \\

&
& TSV
& 61.1 & 60.2 & 59.4 & 65.0 & 55.2 & 52.9 & 59.0 \\

\cmidrule(lr){2-10}

& \multirow{3}{*}{\shortstack{Trajectory-based\\Methods}}
& TraceDet
& 74.1 & 73.9 & 63.7 & 66.1 & 77.1 & 77.2 & 72.0 \\

&
& DynHD
& \underline{86.1}
& \underline{86.7}
& \underline{85.3}
& \underline{84.2}
& \underline{81.3}
& \underline{81.6}
& \underline{84.2} \\

&
& \textbf{\ours}
& \textbf{89.8}
& \textbf{89.1}
& \textbf{88.1}
& \textbf{88.0}
& \textbf{83.4}
& \textbf{85.1}
& \textbf{87.3} \\

\midrule

% ==================== Dream ====================
\multirow{8}{*}{
    \rotatebox[origin=c]{90}{\textbf{Dream-7B-Instruct}}
}
& \multirow{2}{*}{\shortstack{Output-based\\Methods}}
& Semantic Entropy
& 72.5 & 73.7 & 67.7 & 62.7 & 48.6 & 51.4 & 62.8 \\

&
& Lexical Similarity
& 64.0 & 58.3 & 62.7 & 59.7 & 76.9 & 77.3 & 66.5 \\

\cmidrule(lr){2-10}

& \multirow{3}{*}{\shortstack{Latent-based\\Methods}}
& EigenScore
& 69.1 & 66.0 & 67.0 & 62.5 & 77.5 & 76.9 & 69.8 \\

&
& CCS
& 50.3 & 56.9 & 58.2 & 51.7 & 53.2 & 54.2 & 54.1 \\

&
& TSV
& 74.7 & 75.6 & 63.0 & 58.7 & 56.8 & 62.3 & 65.2 \\

\cmidrule(lr){2-10}

& \multirow{3}{*}{\shortstack{Trajectory-based\\Methods}}
& TraceDet
& \underline{86.7}
& 78.1
& 76.0
& 75.1
& 84.1
& \underline{84.7}
& 80.8 \\

&
& DynHD
& 84.4
& \underline{87.3}
& \underline{85.6}
& \underline{80.1}
& \underline{84.6}
& 83.5
& \underline{84.3} \\

&
& \textbf{\ours}
& \textbf{87.7}
& \textbf{88.9}
& \textbf{88.2}
& \textbf{84.6}
& \textbf{86.9}
& \textbf{87.9}
& \textbf{87.4} \\

\bottomrule
\end{tabularx}
}
\vspace{-1mm}
\caption{AUROC (\%) comparison of hallucination detection methods on two D-LLMs across three QA datasets. The highest score is \textbf{bolded}, and the second-highest score is \underline{underlined}.}
\vspace{-1mm}
\label{table:main}
\end{table*}

\noindent\textbf{Variable-Wise Temporal Dynamics Modeling.}
After incorporating stage-adaptive relational evidence into each latent variable, we further model how its dependency-enhanced state evolves throughout the denoising trajectory. 
To capture temporal evolution patterns, we first incorporate trajectory-order information into the dependency-enhanced variable states: 
$\bar{\mathbf{U}}_{k}=\mathbf{U}_{k} + \mathbf{e}^{\mathrm{time}},$
where $\mathbf{e}^{\mathrm{time}}$ denotes the temporal embedding of trajectory step $t$. 
For latent variable $k$, we then apply temporal self-attention to its complete trajectory:
\begin{equation}
\small
\mathbf{R}_{k}
=
\mathrm{TemporalAttn}\left(\bar{\mathbf{U}}_{k}\right).
\label{eq:temporal_attention}
\end{equation}
% This operation captures the evolution pattern of each latent variable throughout the generation trajectory. 
Applying it to all variables yields the multivariate temporal representation $\mathbf{R}$ for hallucination prediction.
Finally, the hallucination probability is predicted through a lightweight classifier:
\(\hat{y} = \operatorname{MLP}(\mathbf{R})\).

\noindent\textbf{Training Objective.}
The detector is trained with a classification objective together with the preservation constraint introduced in T2V. Given the ground-truth hallucination label $y \in \{0,1\}$ and the predicted probability $\hat{y}$, we define the classification loss as the binary cross-entropy:
\begin{equation}
\small
\mathcal{L}_{cls}=-\mathbb{E}\left[y\log \hat{y} +  (1-y)\log(1-\hat{y}) \right].
\label{eq:classification_loss}
\end{equation}
The overall training objective is:
\begin{equation}
\small
\mathcal{L} = \mathcal{L}_{cls} + \mathcal{L}_{pres},
\label{eq:overall_objective}
\end{equation}
which optimizes the detector for hallucination prediction while encouraging the learned variable trajectory to retain fine-grained token-level denoising dynamics. 
Algorithm~\ref{alg:framework} summarizes the procedure of \ours in Appendix~\ref{app:algorithm}, with complexity analysis in Appendix~\ref{complexity}.

\section{Experiments}
\subsection{Experimental Setup}
\noindent\textbf{Datasets.}
Our evaluation covers three complementary question-answering scenarios. \textbf{TriviaQA}~\cite{triviaqa} focuses on open-domain knowledge recall, \textbf{HotpotQA}~\cite{hotpotqa} requires the integration of multiple factual clues, and \textbf{CommonsenseQA}~\cite{commonsenseqa} examines reasoning grounded in everyday knowledge. 
We collect responses and their complete denoising trajectories from two representative D-LLMs, LLaDA-8B-Instruct~\cite{llada} and Dream-7B-Instruct~\cite{dream}, using step generation lengths of 64 and 128.

\noindent\textbf{Baselines and Evaluation.}
We compare \ours with three categories of hallucination detectors:
\ding{182}~\textbf{Output-based methods}, including Perplexity~\cite{perplexity}, Length-Normalized Entropy (LN-Entropy)~\cite{lnentropy}, Semantic Entropy~\cite{semanticentropy}, and Lexical Similarity~\cite{lexicalsimilarity};
\ding{183}~\textbf{Latent-based methods}, including EigenScore~\cite{eigenscore}, Contrast-Consistent Search (CCS)~\cite{ccs}, and Truthfulness Separator Vector (TSV)~\cite{tsv};
\ding{184}~\textbf{Trajectory-based methods}, including TraceDet~\cite{tracedet} and DynHD~\cite{dynhd}. 
We use AUROC as the evaluation metric. Hallucination labels are determined by Qwen3-8B~\cite{qwen3,llm_as_judge_survey} based on the question, answer, and generated response. Detailed descriptions of the baselines and implementation settings are provided in Appendix~\ref{baselines} and Appendix~\ref{implementation}.

\begin{table}[t]
\centering
\scriptsize
\setlength{\tabcolsep}{2.2pt}
\renewcommand{\arraystretch}{1.08}
\resizebox{\columnwidth}{!}{
\begin{tabular}{lccccccc}
\toprule
\multirow{2}{*}{\textbf{Ablation Variant}} & \multicolumn{2}{c}{\textbf{TriviaQA}} & \multicolumn{2}{c}{\textbf{HotpotQA}} & \multicolumn{2}{c}{\textbf{CSQA}} & \multirow{2}{*}{\textbf{Avg.}} \\
\cmidrule(lr){2-3}\cmidrule(lr){4-5}\cmidrule(lr){6-7}
& 64 & 128 & 64 & 128 & 64 & 128 & \\
\midrule
\rowcolor[HTML]{F5F5F5}
\textbf{\ours} & \textbf{89.84} & \textbf{89.13} & \textbf{88.14} & \textbf{88.03} & \textbf{83.42} & \textbf{85.08} & \textbf{87.27} \\
w/o SAUN & 84.32 & 84.72 & 85.65 & 84.37 & 80.05 & 82.05 & 83.53 \\
w/o T2V & 81.78 & 87.40 & 78.60 & 75.65 & 81.15 & 74.04 & 79.77 \\
w/o SAIVI & 83.41 & 86.45 & 81.70 & 81.37 & 79.21 & 74.89 & 81.17 \\
w/o VWTDM & 83.58 & 86.54 & 87.97 & 84.20 & 78.57 & 76.18 & 82.84 \\
\hline
w/o $\mathcal{L}_{amp}$ & 86.59 & 86.97 & 79.97 & 79.85 & 81.53 & 77.53 & 82.07 \\
w/o $\mathcal{L}_{dir}$ & 82.17 & 85.43 & 80.82 & 77.61 & 80.07 & 76.92 & 80.50 \\
\bottomrule
\end{tabular}
}
\vspace{-2mm}
\caption{Ablation study on LLaDA-8B-Instruct.}
\label{tab:ablation}
\vspace{-2mm}
\end{table}

\subsection{Experimental Results and Analysis}
\noindent\textbf{Performance Comparison.} Table~\ref{table:main} presents the main comparison across two D-LLM backbones, three QA benchmarks, and two generation-length settings. We further analyze the experimental results from the perspectives of these three model categories:
\ding{182}
To examine whether final-output statistics are sufficient for D-LLMs hallucination detection, we compare \ours with output-level methods. \ours consistently outperforms output-based baselines across different settings. These results show that output-level evidence cannot adequately characterize hallucinations formed throughout iterative denoising, highlighting the necessity of exploiting denoising-process information.
\ding{183}
To evaluate the benefit of explicitly modeling the generation process over using static internal representations, we compare \ours with latent-based detectors. Although latent-based methods use richer model-side signals than output-level statistics, they still underperform \ours in most cases. This suggests that static representation signals are insufficient to capture the evolving temporal dependencies involved in hallucination formation.
\ding{184}
To verify whether preserving the multivariate structure of denoising trajectories provides additional detection evidence, we compare \ours with TraceDet and DynHD. \ours achieves the best overall performance across different settings. The consistent improvement supports our central motivation: jointly modeling variable-wise evolution and inter-variable interactions better captures abnormal temporal relations for hallucination detection.

\begin{figure}
\vspace{-2mm}
\centering
\includegraphics[width=\linewidth]{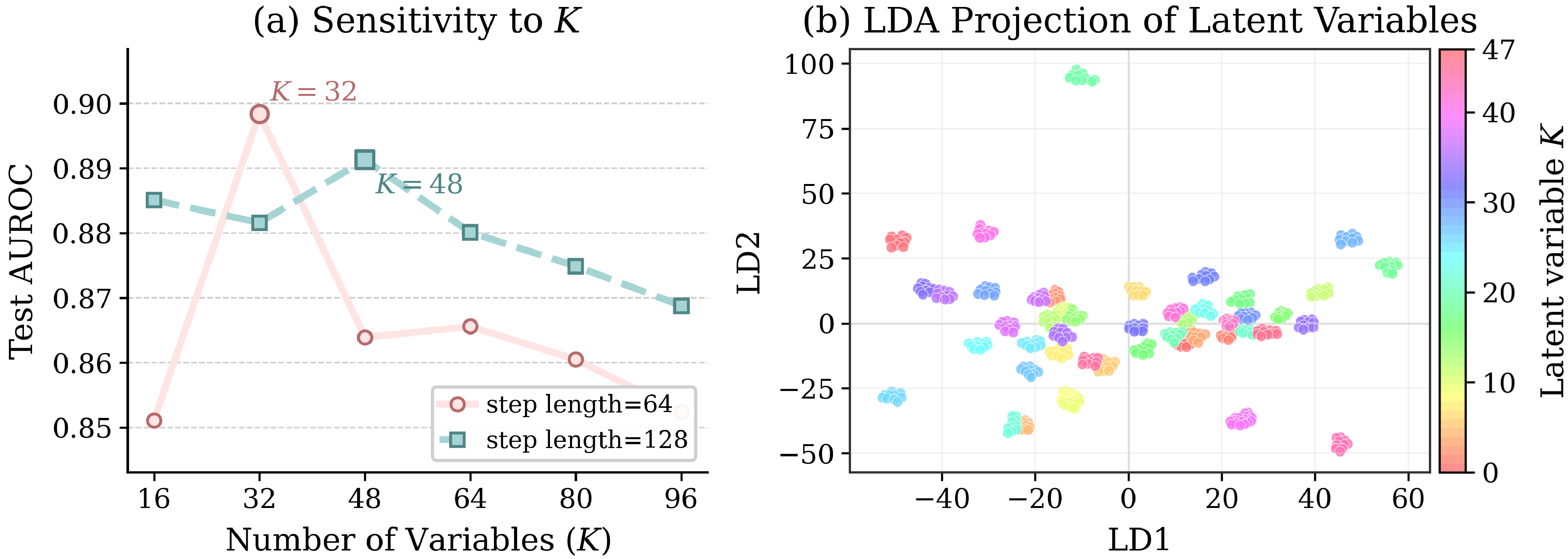}
\vspace{-2mm}
\caption{\textbf{Latent-variable analysis.} Left: performance sensitivity to the number of latent variables (K). Right: visualization of learned variable assignments across different samples.}
\label{figure:visual_K}
\end{figure}

\noindent\textbf{Ablation Study.} 
As shown in Table~\ref{tab:ablation}, removing any component consistently degrades performance on backbones, indicating the effectiveness of multivariate dynamics modeling.
\ding{182} Replacing schedule-normalized signals with raw uncertainty trajectories leads to clear performance drops, suggesting that the schedule-induced global uncertainty trend can mask sample-specific abnormal dynamics. 
\ding{183} Using token positions as variables causes the most pronounced degradation, confirming that semantically unstable token positions do not provide reliable variable identities for time-series modeling. 
\ding{184} Removing either $\mathcal{L}_{amp}$ or $\mathcal{L}_{dir}$ weakens detection performance, demonstrating the importance of preserving fine-grained denoising dynamics during T2V assignment. We provide a sensitivity analysis of the loss hyperparameters in the Appendix~\ref{losssensitivity}.
\ding{185} Removing either inter-variable interaction or temporal evolution modeling also substantially weakens detection performance, showing that trajectory-level hallucination evidence depends on both variable-wise evolution and cross-variable dependency modeling. Comprehensive ablation studies are provided in Appendix~\ref{appendix:complete_ablation}.

\begin{table}[t]
\centering
\scriptsize
\setlength{\tabcolsep}{2.6pt}
\renewcommand{\arraystretch}{1.08}

\resizebox{\columnwidth}{!}{
\begin{tabular}{llccccccc}
\toprule
\multirow{2}{*}{\textbf{Category}}
& \multirow{2}{*}{\textbf{Method}}
& \multicolumn{2}{c}{\textbf{TriviaQA}}
& \multicolumn{2}{c}{\textbf{HotpotQA}}
& \multicolumn{2}{c}{\textbf{CSQA}}
& \multirow{2}{*}{\textbf{Avg.}} \\
\cmidrule(lr){3-4}
\cmidrule(lr){5-6}
\cmidrule(lr){7-8}
& & H-QA & CSQA & T-QA & CSQA & T-QA & H-QA & \\
\midrule

\multirow{2}{*}{\shortstack{\textit{Latent-based}\\\textit{Methods}}}
& CCS
& 50.1 & 54.5 & 51.8 & 54.0 & 54.2 & 56.6 & 53.5 \\

& TSV
& 58.5 & 65.3 & 65.5 & 59.1 & 56.2 & 63.2 & 61.3 \\

\midrule

\multirow{3}{*}{\shortstack{\textit{Trajectory-based}\\\textit{Methods}}}
& TraceDet
& 73.1 & 61.5 & 57.4 & \underline{65.0}
& \textbf{74.8} & 66.2 & 66.3 \\

& DynHD
& \textbf{85.5} & \underline{68.7}
& \underline{73.3} & 64.9
& \underline{73.6} & \underline{71.4}
& \underline{72.9} \\

& \textbf{\ours}
& \underline{82.2} & \textbf{70.2}
& \textbf{74.1} & \textbf{65.8}
& \textbf{74.8} & \textbf{73.6}
& \textbf{73.5} \\

\bottomrule
\end{tabular}
}
\vspace{-1mm}
\caption{Zero-shot cross-task generalization in AUROC (\%).}
\label{table:cross_task}
\end{table}

\noindent\textbf{Robustness to the Number of Latent Variables.} 
As shown in Figure~\ref{figure:visual_K}a, we vary the number of latent variables \(K\) to examine the effect of variable resolution. A small \(K\) over-compresses token-level denoising signals, while a large \(K\) leads to fragmented and redundant variables. The results show that a moderate \(K\) better balances information preservation and variable stability.

\noindent\textbf{Visualization of Learned Variables.} 
To examine whether the Token-to-variable assignment (T2V) module constructs variables with stable identities, we visualize the assignment patterns across multiple samples in Figure~\ref{figure:visual_K}b. Although token contents and semantic roles vary substantially across samples, each latent variable consistently aggregates token-level denoising signals with similar semantic or dynamic characteristics. 
This cross-sample consistency allows the learned latent variables to capture transferable trajectory patterns, thereby improving their generalization to unseen questions and generation contexts.

\noindent\textbf{Cross-Task Generalization.} 
To assess whether \ours captures transferable hallucination dynamics rather than dataset-specific semantic patterns, we train the detector on one benchmark and test it on the other benchmarks without further adaptation. As shown in Table~\ref{table:cross_task}, \ours achieves the highest average AUROC indicating that the stable latent variables constructed through T2V assignment, along with the modeling of temporal evolution and inter-variable interactions, capture transferable hallucination patterns. 

\begin{figure}
\vspace{-2mm}
\centering
\includegraphics[width=0.95\linewidth]{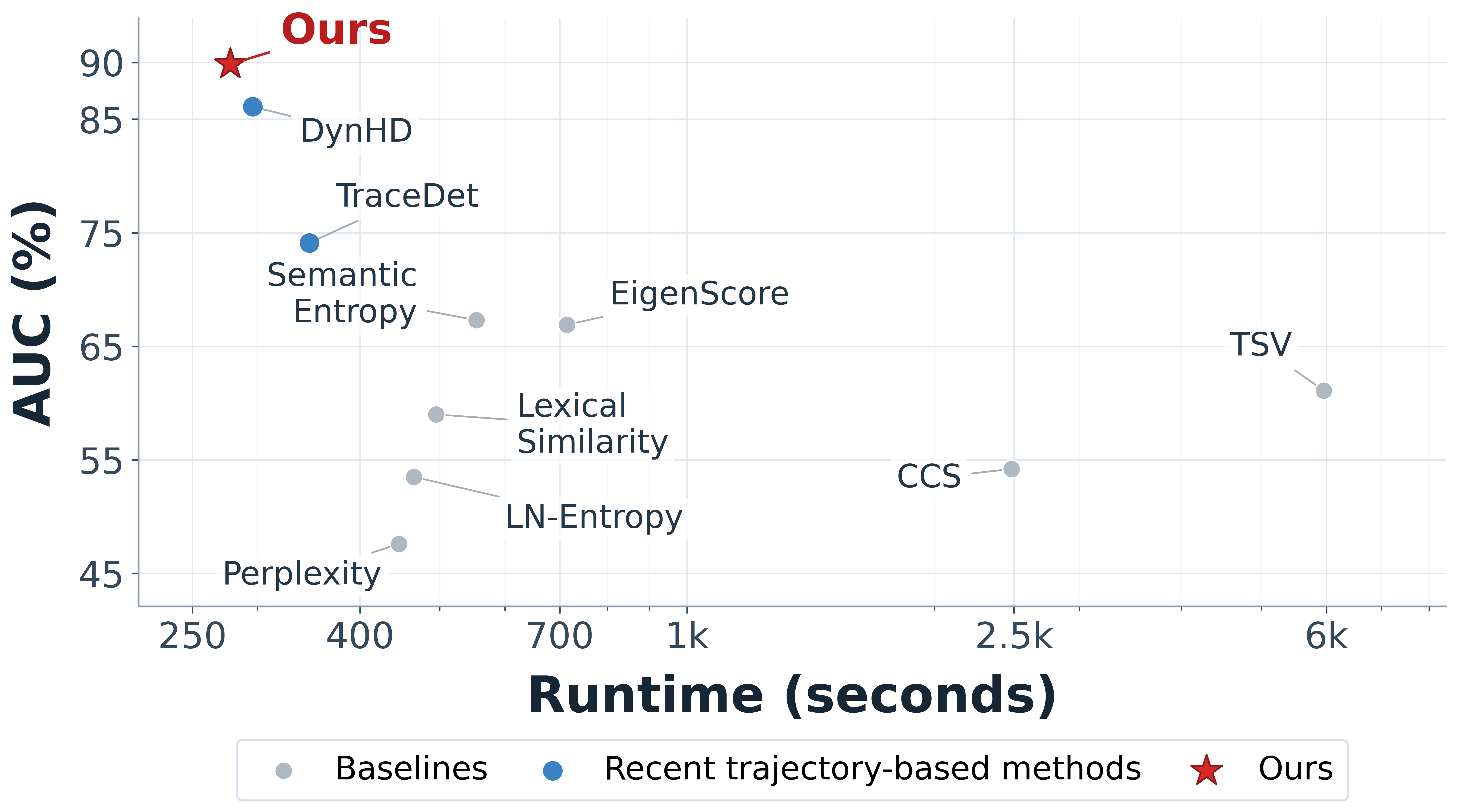}
\vspace{-2mm}
\caption{Average inference time and AUROC of different methods for comprehensive performance evaluation.}
\label{figure:time}
\vspace{-1mm}
\end{figure}

\noindent\textbf{Efficiency Analysis.} 
We compare the inference time and AUROC of different methods to evaluate their practical efficiency as shown in Figure~\ref{figure:time}. \ours achieves the highest AUROC with the shortest runtime. These results show that \ours provides the best AUROC-efficiency trade-off by effectively exploiting multivariate denoising dynamics. 

\section{Conclusion}
In this work, we investigated hallucination detection in D-LLMs from a multivariate time-series perspective. Unlike existing approaches that compress denoising trajectories along the temporal or variable dimension, we preserve their two-dimensional token-step structure to capture how hallucination-related uncertainty evolves and propagates throughout generation. 
Extensive experiments across multiple D-LLMs, benchmarks, and generation settings demonstrate that \ours exhibits strong effectiveness, cross-task generalization, robustness, and computational efficiency.

\clearpage

\bibliography{aaai2027}

@article{shen2026raising,
  title={Raising the bar in graph ood generalization: Invariant learning beyond explicit environment modeling},
  author={Shen, Xu and Liu, Yixin and Wang, Yili and Miao, Rui and Dai, Yiwei and Pan, Shirui and Chang, Yi and Wang, Xin},
  journal={IEEE Transactions on Pattern Analysis and Machine Intelligence},
  year={2026}
}

@inproceedings{zhao2026fedcigar,
  title={FedCIGAR: A Personalized Reconstruction Approach for Federated Graph-level Anomaly Detection},
  author={Zhao, Yunfeng and Liu, Yixin and Chen, Qingfeng and Li, Shiyuan and Tan, Yue and Pan, Shirui},
  booktitle={International Joint Conference on Artificial Intelligence},
  year={2026}
}

@article{li2026towards,
  title={Towards Anomaly Detection on Relational Data},
  author={Li, Shiyuan and Zhao, Yunfeng and Tan, Yue and Chen, Qingfeng and Liu, Yixin and Pan, Shirui},
  journal={arXiv preprint arXiv:2606.18621},
  year={2026}
}

@article{chen2025multi,
  title={Multi-Stage Verification-Centric Framework for Mitigating Hallucination in Multi-Modal RAG},
  author={Chen, Baiyu and Wongso, Wilson and Hu, Xiaoqian and Tan, Yue and Salim, Flora},
  journal={arXiv preprint arXiv:2507.20136},
  year={2025}
}

@article{tan2024influence,
  title={Influence-oriented personalized federated learning},
  author={Tan, Yue and Long, Guodong and Jiang, Jing and Zhang, Chengqi},
  journal={arXiv preprint arXiv:2410.03315},
  year={2024}
}

@inproceedings{pan2026explainable,
  title={Explainable and Fine-Grained Safeguarding of LLM Multi-Agent Systems via Bi-Level Graph Anomaly Detection},
  author={Pan, Junjun and Liu, Yixin and Miao, Rui and Ding, Kaize and Zheng, Yu and Nguyen, Quoc Viet Hung and Liew, Alan Wee-Chung and Pan, Shirui},
  booktitle={Proceedings of the 64th Annual Meeting of the Association for Computational Linguistics},
  year={2026}
}

@inproceedings{he2026dual,
  title={Dual Mamba for Node-Specific Representation Learning: Tackling Over-Smoothing with Selective State Space Modeling},
  author={He, Xin and Wang, Yili and Dai, Yiwei and Wang, Xin},
  booktitle={Proceedings of the AAAI Conference on Artificial Intelligence},
  volume={40},
  number={26},
  pages={21672--21680},
  year={2026}
}

@article{bhatnagar2026drift,
  title={DRIFT: Detecting Representational Inconsistencies for Factual Truthfulness},
  author={Bhatnagar, Rohan and Sun, Youran and Zhang, Chi Andrew and Wen, Yixin and Yang, Haizhao},
  journal={arXiv e-prints},
  pages={arXiv--2601},
  year={2026}
}

@article{badave2026beyond,
  title={Beyond Final Answers: Auditing Trajectory-Level Hallucinations in Multi-Agent Industrial Workflows},
  author={Badave, Harshada and Borse, Santosh and Gomez, Andrea and Narahari, Harshitha and Carter, Sara and Bhatt, Vishwa and Rachakonda, Aishani and Lin, Shuxin and Patel, Dhaval},
  journal={arXiv preprint arXiv:2605.24219},
  year={2026}
}

@inproceedings{shoby2026overthinking,
  title={Overthinking Causes Hallucination: Tracing Confounder Propagation in Vision Language Models},
  author={Shoby, Abin and Huy, Ta Duc and Nguyen, Tuan Dung and Ho, Minh Khoi and Chen, Qi and van den Hengel, Anton and Le Nguyen, Phi and Verjans, Johan W and Phan, Vu Minh Hieu},
  booktitle={Proceedings of the IEEE/CVF Conference on Computer Vision and Pattern Recognition},
  pages={9185--9194},
  year={2026}
}

@inproceedings{zhang2025icr,
  title={ICR probe: Tracking hidden state dynamics for reliable hallucination detection in LLMs},
  author={Zhang, Zhenliang and Hu, Xinyu and Zhang, Huixuan and Zhang, Junzhe and Wan, Xiaojun},
  booktitle={Proceedings of the 63rd Annual Meeting of the Association for Computational Linguistics (Volume 1: Long Papers)},
  pages={17986--18002},
  year={2025}
}

@article{chen2026hallusae,
  title={HalluSAE: Detecting Hallucinations in Large Language Models via Sparse Auto-Encoders},
  author={Chen, Boshui and Fan, Zhaoxin and Wang, Ke and Leng, Zhiying and Wu, Faguo and Zheng, Hongwei and Sun, Yifan and Wu, Wenjun},
  journal={arXiv preprint arXiv:2604.16430},
  year={2026}
}

@article{weng2026tre,
  title={TRE: Training-Free Hallucination Detection for Diffusion Language Models},
  author={Weng, Pengcheng and Qian, Yanyu and Tan, Yue and Liu, Yixin},
  year={2026},
  publisher={Preprints}
}

@article{mehri2018middle,
  title={Middle-out decoding},
  author={Mehri, Shikib and Sigal, Leonid},
  journal={Advances in Neural Information Processing Systems},
  volume={31},
  year={2018}
}

@article{gu2019insertion,
  title={Insertion-based decoding with automatically inferred generation order},
  author={Gu, Jiatao and Liu, Qi and Cho, Kyunghyun},
  journal={Transactions of the Association for Computational Linguistics},
  volume={7},
  pages={661--676},
  year={2019},
  publisher={MIT Press One Rogers Street, Cambridge, MA 02142-1209, USA journals-info~…}
}

@inproceedings{ghazvininejad2019mask,
  title={Mask-predict: Parallel decoding of conditional masked language models},
  author={Ghazvininejad, Marjan and Levy, Omer and Liu, Yinhan and Zettlemoyer, Luke},
  booktitle={Proceedings of the 2019 conference on empirical methods in natural language processing and the 9th international joint conference on natural language processing (EMNLP-IJCNLP)},
  pages={6112--6121},
  year={2019}
}

@article{vcformer,
  title={Vcformer: Variable correlation transformer with inherent lagged correlation for multivariate time series forecasting},
  author={Yang, Yingnan and Zhu, Qingling and Chen, Jianyong},
  journal={arXiv preprint arXiv:2405.11470},
  year={2024}
}

@article{tqnet,
  title={Temporal query network for efficient multivariate time series forecasting},
  author={Lin, Shengsheng and Chen, Haojun and Wu, Haijie and Qiu, Chunyun and Lin, Weiwei},
  journal={arXiv preprint arXiv:2505.12917},
  year={2025}
}

@article{diffusionmodel,
  title={Diffusion models: A comprehensive survey of methods and applications},
  author={Yang, Ling and Zhang, Zhilong and Song, Yang and Hong, Shenda and Xu, Runsheng and Zhao, Yue and Zhang, Wentao and Cui, Bin and Yang, Ming-Hsuan},
  journal={ACM computing surveys},
  volume={56},
  number={4},
  pages={1--39},
  year={2023},
  publisher={ACM New York, NY, USA}
}

@article{llm_as_judge_survey,
  title={A survey on llm-as-a-judge},
  author={Gu, Jiawei and Jiang, Xuhui and Shi, Zhichao and Tan, Hexiang and Zhai, Xuehao and Xu, Chengjin and Li, Wei and Shen, Yinghan and Ma, Shengjie and Liu, Honghao and others},
  journal={The Innovation},
  volume={7},
  number={6},
  year={2026},
  publisher={Elsevier}
}

@article{qwen3,
  title={Qwen3 technical report},
  author={Yang, An and Li, Anfeng and Yang, Baosong and Zhang, Beichen and Hui, Binyuan and Zheng, Bo and Yu, Bowen and Gao, Chang and Huang, Chengen and Lv, Chenxu and others},
  journal={arXiv preprint arXiv:2505.09388},
  year={2025}
}

@article{tsv,
  title={Steer llm latents for hallucination detection},
  author={Park, Seongheon and Du, Xuefeng and Yeh, Min-Hsuan and Wang, Haobo and Li, Yixuan},
  journal={arXiv preprint arXiv:2503.01917},
  year={2025}
}

@article{ccs,
  title={Discovering latent knowledge in language models without supervision},
  author={Burns, Collin and Ye, Haotian and Klein, Dan and Steinhardt, Jacob},
  journal={arXiv preprint arXiv:2212.03827},
  year={2022}
}

@article{eigenscore,
  title={FacTool: Factuality Detection in Generative AI--A Tool Augmented Framework for Multi-Task and Multi-Domain Scenarios},
  author={Chern, I and Chern, Steffi and Chen, Shiqi and Yuan, Weizhe and Feng, Kehua and Zhou, Chunting and He, Junxian and Neubig, Graham and Liu, Pengfei and others},
  journal={arXiv preprint arXiv:2307.13528},
  year={2023}
}

@article{lexicalsimilarity,
  title={Generating with confidence: Uncertainty quantification for black-box large language models, 2024},
  author={Lin, Zhen and Trivedi, Shubhendu and Sun, Jimeng},
  journal={URL https://arxiv. org/abs/2305},
  volume={19187},
  year={2024}
}

@article{semanticentropy,
  title={Semantic uncertainty: Linguistic invariances for uncertainty estimation in natural language generation},
  author={Kuhn, Lorenz and Gal, Yarin and Farquhar, Sebastian},
  journal={arXiv preprint arXiv:2302.09664},
  year={2023}
}

@article{lnentropy,
  title={Uncertainty estimation in autoregressive structured prediction},
  author={Malinin, Andrey and Gales, Mark},
  journal={arXiv preprint arXiv:2002.07650},
  year={2020}
}

@article{perplexity,
  title={Out-of-distribution detection and selective generation for conditional language models},
  author={Ren, Jie and Luo, Jiaming and Zhao, Yao and Krishna, Kundan and Saleh, Mohammad and Lakshminarayanan, Balaji and Liu, Peter J},
  journal={arXiv preprint arXiv:2209.15558},
  year={2022}
}

@inproceedings{commonsenseqa,
  title={Commonsenseqa: A question answering challenge targeting commonsense knowledge},
  author={Talmor, Alon and Herzig, Jonathan and Lourie, Nicholas and Berant, Jonathan},
  booktitle={Proceedings of the 2019 Conference of the North American Chapter of the Association for Computational Linguistics: Human Language Technologies, Volume 1 (Long and Short Papers)},
  pages={4149--4158},
  year={2019}
}

@inproceedings{hotpotqa,
  title={HotpotQA: A dataset for diverse, explainable multi-hop question answering},
  author={Yang, Zhilin and Qi, Peng and Zhang, Saizheng and Bengio, Yoshua and Cohen, William and Salakhutdinov, Ruslan and Manning, Christopher D},
  booktitle={Proceedings of the 2018 conference on empirical methods in natural language processing},
  pages={2369--2380},
  year={2018}
}

@inproceedings{triviaqa,
  title={Triviaqa: A large scale distantly supervised challenge dataset for reading comprehension},
  author={Joshi, Mandar and Choi, Eunsol and Weld, Daniel S and Zettlemoyer, Luke},
  booktitle={Proceedings of the 55th Annual Meeting of the Association for Computational Linguistics (Volume 1: Long Papers)},
  pages={1601--1611},
  year={2017}
}

@article{mtscp3,
  title={Crossgnn: Confronting noisy multivariate time series via cross interaction refinement},
  author={Huang, Qihe and Shen, Lei and Zhang, Ruixin and Ding, Shouhong and Wang, Binwu and Zhou, Zhengyang and Wang, Yang},
  journal={Advances in Neural Information Processing Systems},
  volume={36},
  pages={46885--46902},
  year={2023}
}

@article{mtscp2,
  title={Multiple time series forecasting with dynamic graph modeling},
  author={Zhao, Kai and Guo, Chenjuan and Cheng, Yunyao and Han, Peng and Zhang, Miao and Yang, Bin},
  journal={Proceedings of the VLDB Endowment},
  volume={17},
  number={4},
  pages={753--765},
  year={2023},
  publisher={VLDB Endowment}
}

@inproceedings{mtscp1,
  title={Duet: Dual clustering enhanced multivariate time series forecasting},
  author={Qiu, Xiangfei and Wu, Xingjian and Lin, Yan and Guo, Chenjuan and Hu, Jilin and Yang, Bin},
  booktitle={Proceedings of the 31st ACM SIGKDD Conference on Knowledge Discovery and Data Mining V. 1},
  pages={1185--1196},
  year={2025}
}

@inproceedings{mtsci2,
  title={Are transformers effective for time series forecasting?},
  author={Zeng, Ailing and Chen, Muxi and Zhang, Lei and Xu, Qiang},
  booktitle={Proceedings of the AAAI conference on artificial intelligence},
  volume={37},
  number={9},
  pages={11121--11128},
  year={2023}
}

@article{mtsci1,
  title={A time series is worth 64 words: Long-term forecasting with transformers},
  author={Nie, Yuqi and Nguyen, Nam H and Sinthong, Phanwadee and Kalagnanam, Jayant},
  journal={arXiv preprint arXiv:2211.14730},
  year={2022}
}

@inproceedings{mtscd1,
  title={itransformer: Inverted transformers are effective for time series forecasting},
  author={Liu, Yong and Hu, Tengge and Zhang, Haoran and Wu, Haixu and Wang, Shiyu and Ma, Lintao and Long, Mingsheng},
  booktitle={International conference on learning representations},
  volume={2024},
  pages={11116--11140},
  year={2024}
}

@article{mtssurvey3,
  title={A Survey on Data Generation for Time Series: Taxonomy, Review and Prospects},
  author={Zhang, Xu and Xu, Chang and Li, Hao and Huang, Yuhao and Xu, Qiushui and Liang, Yuxuan and Liu, Chenghao and Jin, Ming and Wen, Qingsong and Wang, Peng and others},
  year={2026},
  publisher={Preprints}
}

@article{mtssurvey2,
  title={Mamba for Time Series Analysis: A Contemporary Survey},
  author={Nguyen, Thanh Tam and Jin, Ming and Pham, Trinh and Pan, Shirui and Nguyen, Quoc Viet Hung},
  year={2026},
  publisher={Preprints}
}

@article{mts1,
  title={Segrnn: Segment recurrent neural network for long-term time series forecasting},
  author={Lin, Shengsheng and Lin, Weiwei and Wu, Wentai and Zhao, Feiyu and Mo, Ruichao and Zhang, Haotong},
  journal={IEEE Internet of Things Journal},
  year={2025},
  publisher={IEEE}
}

@article{llada,
  title={Large language diffusion models},
  author={Nie, Shen and Zhu, Fengqi and You, Zebin and Zhang, Xiaolu and Ou, Jingyang and Hu, Jun and Zhou, Jun and Lin, Yankai and Wen, Ji-Rong and Li, Chongxuan},
  journal={Advances in Neural Information Processing Systems},
  volume={38},
  pages={50608--50646},
  year={2026}
}

@article{dream,
  title={Dream 7b: Diffusion large language models},
  author={Ye, Jiacheng and Xie, Zhihui and Zheng, Lin and Gao, Jiahui and Wu, Zirui and Jiang, Xin and Li, Zhenguo and Kong, Lingpeng},
  journal={arXiv preprint arXiv:2508.15487},
  year={2025}
}

@article{llada100b,
  title={Llada2. 0: Scaling up diffusion language models to 100b},
  author={Bie, Tiwei and Cao, Maosong and Chen, Kun and Du, Lun and Gong, Mingliang and Gong, Zhuochen and Gu, Yanmei and Hu, Jiaqi and Huang, Zenan and Lan, Zhenzhong and others},
  journal={arXiv preprint arXiv:2512.15745},
  year={2025}
}

@inproceedings{hallucinationofDLLM1,
  title={Towards understanding text hallucination of diffusion models via local generation bias},
  author={Lu, Rui and Wang, Runzhe and Lyu, Kaifeng and Jiang, Xitai and Huang, Gao and Wang, Mengdi},
  booktitle={The Thirteenth International Conference on Learning Representations},
  year={2025}
}

@article{hallucinationofDLLM2,
  title={From Denoising to Refining: A Corrective Framework for Vision-Language Diffusion Model},
  author={Ji, Yatai and Wang, Teng and Ge, Yuying and Liu, Zhiheng and Yang, Sidi and Shan, Ying and Luo, Ping},
  journal={arXiv preprint arXiv:2510.19871},
  year={2025}
}

@inproceedings{hallucination,
  title={Hallulens: Llm hallucination benchmark},
  author={Bang, Yejin and Ji, Ziwei and Schelten, Alan and Hartshorn, Anthony and Fowler, Tara and Zhang, Cheng and Cancedda, Nicola and Fung, Pascale},
  booktitle={Proceedings of the 63rd Annual Meeting of the Association for Computational Linguistics (Volume 1: Long Papers)},
  pages={24128--24156},
  year={2025}
}

@article{hallucination2,
  title={Why language models hallucinate},
  author={Kalai, Adam Tauman and Nachum, Ofir and Vempala, Santosh S and Zhang, Edwin},
  journal={arXiv preprint arXiv:2509.04664},
  year={2025}
}

@article{tdgnet,
  title={TDGNet: Hallucination Detection in Diffusion Language Models via Temporal Dynamic Graphs},
  author={Hemmat, Arshia and Torr, Philip and Chen, Yongqiang and Yu, Junchi},
  journal={arXiv preprint arXiv:2602.08048},
  year={2026}
}

@article{tracedet,
  title={TraceDet: Hallucination Detection from the Decoding Trace of Diffusion Large Language Models},
  author={Chang, Shenxu and Yu, Junchi and Wang, Weixing and Chen, Yongqiang and Yu, Jialin and Torr, Philip and Gu, Jindong},
  journal={arXiv preprint arXiv:2510.01274},
  year={2025}
}

@article{hive,
  title={HIVE: Hidden-Evidence Verification for Hallucination Detection in Diffusion Large Language Models},
  author={Zhao, Guoshenghui and Zhao, Weijie and Yu, Tan},
  journal={arXiv preprint arXiv:2604.26139},
  year={2026}
}

@article{hallucinationofDLLM3,
  title={Lost in Diffusion: Uncovering Hallucination Patterns and Failure Modes in Diffusion Large Language Models},
  author={Guo, Zhengnan and Tan, Fei},
  journal={arXiv preprint arXiv:2604.10556},
  year={2026}
}

@article{dynhd,
  title={Dynhd: Hallucination detection for diffusion large language models via denoising dynamics deviation learning},
  author={Qian, Yanyu and Tan, Yue and Liu, Yixin and Yu, Wang and Pan, Shirui},
  journal={arXiv preprint arXiv:2603.16459},
  year={2026}
}

@article{mtssurvey,
  title={A comprehensive survey of deep learning for multivariate time series forecasting: A channel strategy perspective},
  author={Qiu, Xiangfei and Cheng, Hanyin and Wu, Xingjian and Lu, Junkai and Hu, Jilin and Guo, Chenjuan and Jensen, Christian S and Yang, Bin},
  journal={arXiv preprint arXiv:2502.10721},
  year={2025}
}

@article{timefilter,
  title={TimeFilter: Patch-specific spatial-temporal graph filtration for time series forecasting},
  author={Hu, Yifan and Zhang, Guibin and Liu, Peiyuan and Lan, Disen and Li, Naiqi and Cheng, Dawei and Dai, Tao and Xia, Shu-Tao and Pan, Shirui},
  journal={arXiv preprint arXiv:2501.13041},
  year={2025}
}

@inproceedings{crossformer,
  title={Crossformer: Transformer utilizing cross-dimension dependency for multivariate time series forecasting},
  author={Zhang, Yunhao and Yan, Junchi},
  booktitle={The eleventh international conference on learning representations},
  year={2023}
}
\clearpage
\appendix
\setcounter{secnumdepth}{2} \setcounter{section}{0} \renewcommand{\thesection}{\Alph{section}} \renewcommand{\thesubsection}{\Alph{section}.\arabic{subsection}} \renewcommand{\thesubsubsection}{\Alph{section}.\arabic{subsection}.\arabic{subsubsection}}

\section{Extended Related Work}\label{app:rw}
\subsection{Hallucination Detection in D-LLMs}

D-LLMs generate text through iterative denoising, which exposes intermediate predictions, uncertainty estimates, and hidden states before the final response is obtained~\cite{diffusionmodel,llada,dream}. This process provides trajectory-level evidence for hallucination detection, since factual errors may emerge during refinement rather than only in the final output~\cite{hallucinationofDLLM3}. Existing studies mainly explore this evidence from two perspectives.

The first line focuses on \textit{step-oriented evidence selection}. TraceDet~\cite{tracedet} formulates the denoising process as a decoding trace and selects informative sub-traces for hallucination prediction, showing that different denoising stages contribute unequally to detection. HIVE~\cite{hive} further exploits hidden evidence from intermediate denoising states and uses selected evidence for verification. These methods demonstrate the value of intermediate states, but they mainly rely on selecting or compressing denoising stages.

The second line focuses on \textit{dynamics-oriented trajectory modeling}. DynHD~\cite{dynhd} models the evolution of uncertainty evidence and detects hallucinations by measuring deviations from expected denoising dynamics. TDGNet~\cite{tdgnet} constructs temporal dynamic graphs to capture evolving token-level relations during denoising. These methods reveal that hallucination cues can lie in dynamic uncertainty patterns and token interactions. However, they still do not fully preserve the complete token-by-step trajectory structure. In contrast, \ours models D-LLM denoising trajectories as multivariate time series, aiming to jointly capture token-wise temporal evolution and evolving inter-token dependencies for hallucination detection.

\subsection{Multivariate Time-Series Modeling}

Multivariate time-series modeling aims to capture temporal patterns across multiple variables and their dependencies~\cite{mts1,li2026towards,mtssurvey2}. Existing methods are often discussed according to how they model variable channels. Channel-independent methods, such as PatchTST~\cite{mtsci1}, model each variable separately to preserve variable-specific temporal patterns and reduce interference from noisy correlations. Channel-dependent methods, such as Crossformer~\cite{crossformer} and iTransformer~\cite{mtscd1}, jointly model variables to capture cross-variable dependencies. More recent channel-partial methods, such as DUET~\cite{mtscp1}, CrossGNN~\cite{mtscp3}, and TimeFilter~\cite{timefilter}, allow each variable to interact with only relevant variables, balancing variable-specific modeling and dependency modeling.

These methods provide useful inspiration for modeling D-LLMs denoising trajectories~\cite{chen2025multi,tan2024influence,pan2026explainable,he2026dual}, where denoising steps form the temporal axis and token signals form the raw variable axis. However, conventional multivariate time-series methods usually assume that each variable has a stable identity and meaning across samples and time. This assumption does not hold for D-LLMs, because token positions may correspond to different semantic roles across samples and may change during denoising. Therefore, directly treating token positions as variables can lead to unstable temporal patterns and unreliable dependencies. To address this issue, \ours first converts changing token signals into stable latent variables, and then models both their temporal evolution and stage-adaptive inter-variable interactions.

\section{Algorithm}\label{app:algorithm}
For completeness, we summarize the overall workflow and optimization procedure of the proposed framework in Algorithm~\ref{alg:framework}, providing a concise overview of its implementation. 

\begin{algorithm}
\caption{Overall Procedure of \ours}
\label{alg:framework}
\begin{algorithmic}[1]
\STATE \textbf{Input:} Token uncertainty trajectory $\mathbf{H}\in\mathbb{R}^{T\times N}$, contextual token embeddings $\{\mathbf{e}_{i,t}\}$, label $y$

\STATE \textbf{Output:} Hallucination probability $\hat{y}$ and objective $\mathcal{L}$

\STATE // \textbf{Module 1: Trajectory-Preserving Token-to-Variable Assignment (T2V)}

\STATE Normalize token uncertainty $\widehat{\mathbf{H}}$ via Eq.~\eqref{eq:normalization}.

\FOR{each denoising step $t$ and token position $i$}
    \STATE Compute uncertainty-semantic representation $\mathbf{u}_{i,t}$ via Eq.~\eqref{eq:he}.
    \STATE Compute content assignment logits $\mathbf{s}^{\mathrm{con}}_{i,t}$ via Eq.~\eqref{eq:content_assignment_logits}.
    \STATE Compute positional representation $\boldsymbol{\phi}_{i}$ and structural logits $\mathbf{s}^{\mathrm{str}}_{i}$ via Eqs.~\eqref{eq:position}--\eqref{eq:structural_assignment_logits}.
    \STATE Compute gate $\mathbf{g}_{i,t}$ and assignment weights $\mathbf{a}_{i,t}$ via Eqs.~\eqref{eq:content_adaptive_gate}--\eqref{eq:assignment}.
\ENDFOR

\FOR{each denoising step $t$ and latent variable $k$}
    \STATE Construct latent-variable state $\mathbf{z}_{k,t}=\mathcal{A}_{k,t}(\mathbf{u}_{t})$ via Eq.~\eqref{eq:aggregation_operator}.
\ENDFOR

\STATE Form latent-variable trajectory $\mathbf{Z}=\{\mathbf{z}_{k,t}\}_{k=1,t=1}^{K,T}$.
\STATE Compute preservation loss $\mathcal{L}_{pres}$ via Eqs.~\eqref{eq:amplitude_preservation}--\eqref{eq:preservation_constraint}.

\STATE // \textbf{Module 2: Dynamic Multivariate Temporal Modeling (DMTM)}
\FOR{each denoising step $t$}
    \STATE Compute inter-variable interaction $\mathbf{C}_{t}$ via Eq.~\eqref{eq:inter_variable_interaction}.
    \STATE Compute reliability-gated variable states $\mathbf{U}_{t}$ via Eq.~\eqref{eq:featuremerge}.
\ENDFOR

\FOR{each latent variable $k$}
    \STATE Encode variable-wise temporal dynamics $\mathbf{R}_{k}$ via Eq.~\eqref{eq:temporal_attention}.
\ENDFOR

\STATE Aggregate $\{\mathbf{R}_{k}\}_{k=1}^{K}$ into $\mathbf{R}$ and predict $\hat{y}=\operatorname{MLP}(\mathbf{R})$.
\STATE Compute $\mathcal{L}_{cls}$ and the overall objective $\mathcal{L}$ via Eqs.~\eqref{eq:classification_loss}--\eqref{eq:overall_objective}.

\RETURN $\hat{y}$ and $\mathcal{L}$.
\end{algorithmic}
\end{algorithm}

\section{Computational Complexity}\label{complexity}

Let \(T\), \(N\), and \(K\) denote the numbers of denoising steps, token positions, and latent variables, respectively, and let \(d=d_u\) denote the dimension of the token and latent-variable representations. Following the standard convention for attention complexity, we report the dominant assignment, aggregation, and pairwise interaction costs introduced by \ours, excluding the original D-LLM denoising process, token-wise transformations, and the lightweight classifier.

In T2V, SAUN normalizes the uncertainty distribution over all token positions at each denoising step, resulting in \(\mathcal{O}(TN)\) complexity. The subsequent token-to-variable assignment computes the assignment weights \(a_{i,t,k}\), while assignment-weighted aggregation constructs the latent-variable states \(\mathbf{z}_{k,t}\) from the token representations. These operations require \(\mathcal{O}(TNKd)\) computation. During training, the trajectory-preservation constraint aggregates the \(d\)-dimensional token-level variations \(\Delta\widehat{\mathbf H}_{i,t}\) into \(\Delta\mathbf Z_{k,t}\) and reconstructs \(\Delta\widetilde{\mathbf H}_{i,t}\), introducing an additional \(\mathcal{O}(TNKd)\) cost. This constraint incurs no additional cost at inference.

In DMTM, SAIVI models interactions among \(K\) latent variables independently at each denoising step, resulting in a pairwise attention cost of \(\mathcal{O}(TK^{2}d)\). VTDM performs temporal self-attention over \(T\) denoising steps for each latent variable, requiring \(\mathcal{O}(KT^{2}d)\) computation. Therefore, the overall inference-time complexity is

\begin{equation}\mathcal{O}\left(TNKd+TK^{2}d+KT^{2}d\right).\label{eq:computational_complexity}\end{equation}

During training, the trajectory-preservation constraint introduces an additional \(\mathcal{O}(TNKd)\) term, which has the same asymptotic order as the token-to-variable assignment and aggregation.

The main efficiency advantage of \ours arises from conducting relational and temporal modeling in the compact latent-variable space. Applying analogous axial attention directly to the original token-indexed trajectory would model interactions among \(N\) token positions at each denoising step and temporal dependencies over \(T\) steps for each token position, requiring \(\mathcal{O}(TN^{2}d+NT^{2}d)\) attention computation. In contrast, \ours first transforms the token-indexed trajectory into \(K\) latent-variable trajectories with an assignment and aggregation cost of \(\mathcal{O}(TNKd)\), and reduces the subsequent attention cost to \(\mathcal{O}(TK^{2}d+KT^{2}d)\). Since \(K< N\) in practice, this design reduces the quadratic relational cost from \(N^{2}\) to \(K^{2}\) and decreases the number of temporally modeled trajectories from \(N\) to \(K\). Moreover, \ours directly reuses the uncertainty and contextual signals collected during denoising, requiring neither repeated generation nor additional forward passes through the D-LLM backbone.

\begin{table}[t]
\centering
\scriptsize
\setlength{\tabcolsep}{2.2pt}
\renewcommand{\arraystretch}{1.08}

\resizebox{\columnwidth}{!}{
\begin{tabular}{lcccccc}
\toprule
\multirow{2}{*}{\textbf{Ablation Variant}}
& \multicolumn{2}{c}{\textbf{TriviaQA}}
& \multicolumn{2}{c}{\textbf{HotpotQA}}
& \multicolumn{2}{c}{\textbf{CSQA}} \\
\cmidrule(lr){2-3}
\cmidrule(lr){4-5}
\cmidrule(lr){6-7}
& 64 & 128 & 64 & 128 & 64 & 128 \\
\midrule

\rowcolor[HTML]{DEDEDE}
\multicolumn{7}{l}{\textbf{LLaDA-8B-Instruct}} \\

\rowcolor[HTML]{F5F5F5}
\textbf{\ours}
& \textbf{89.84}
& \textbf{89.13}
& \textbf{88.14}
& \textbf{88.03}
& \textbf{83.42}
& \textbf{85.08} \\

% w/o Schedule-Aware Uncertainty Normalization
w/o SAUN
& 84.32 & 84.72 & 85.65 & 84.37 & 80.05 & 82.05 \\

w/o T2V
& 81.78 & 87.40 & 78.60 & 75.65 & 81.15 & 74.04 \\

% w/o Stage-Adaptive Inter-Variable Interaction 
w/o SAIVI
& 83.41 & 86.45 & 81.70 & 81.37 & 79.21 & 74.89 \\

% w/o Variable-wise Temporal Dynamics Modeling
w/o VWTDM
& 83.58 & 86.54 & 87.97 & 84.20 & 78.57 & 76.18 \\

w/o $\mathcal{L}_{amp}$
& 86.59 & 86.97 & 79.97 & 79.85 & 81.53 & 77.53 \\

w/o $\mathcal{L}_{dir}$ 
& 82.17 & 85.43 & 80.82 & 77.61 & 80.07 & 76.92 \\

\midrule

\rowcolor[HTML]{DEDEDE}
\multicolumn{7}{l}{\textbf{Dream-7B-Instruct}} \\

\rowcolor[HTML]{F5F5F5}
\textbf{\ours}
& \textbf{87.74}
& \textbf{88.89}
& \textbf{88.16}
& \textbf{84.63}
& \textbf{86.95}
& \textbf{87.86} \\

% w/o Schedule-Aware Uncertainty Normalization
w/o SAUN
& 81.07 & 82.90 & 79.50 & 79.04 & 83.07 & 86.67 \\

w/o T2V
& 80.50 & 84.84 & 79.17 & 77.84 & 77.18 & 83.85 \\

% w/o Stage-Adaptive Inter-Variable Interaction 
w/o SAIVI
& 77.95 & 86.11 & 79.87 & 72.77 & 81.38 & 83.01 \\

% w/o Variable-wise Temporal Dynamics Modeling
w/o VWTDM
& 80.27 & 82.32 & 80.98 & 74.65 & 82.61 & 84.21 \\

w/o $\mathcal{L}_{amp}$
& 79.61 & 82.17 & 81.21 & 78.47 & 78.57 & 83.41 \\

w/o $\mathcal{L}_{dir}$ 
& 82.13 & 81.29 & 80.15 & 79.96 & 78.69 & 82.06 \\
\bottomrule
\end{tabular}
}
\caption{Ablation study in AUROC (\%) on two D-LLMs backbones.}
\label{tab:full_ablation}
\end{table}

\section{Experiments}

\subsection{Baselines}\label{baselines}
We compare \ours with three categories of hallucination detectors, covering output-level signals, internal representations, and denoising-trajectory evidence. 

\ding{182}\textbf{Output-based methods.} These methods assess hallucination risk using information derived from the generated responses, including model likelihood, token-level uncertainty, and consistency among sampled outputs. They serve as widely used baselines for evaluating whether final-output signals are sufficient for hallucination detection.

\noindent$\rhd$\textbf{Perplexity}~\cite{perplexity} uses the likelihood of the generated response as a confidence signal, where responses with lower model likelihood are considered less reliable.

\noindent$\rhd$\textbf{Length-Normalized Entropy (LN-Entropy)}~\cite{lnentropy} estimates predictive uncertainty from token-level entropy and normalizes it by sequence length, reducing the bias caused by different output lengths.

\noindent$\rhd$\textbf{Semantic Entropy}~\cite{semanticentropy} samples multiple responses, groups them according to semantic equivalence, and computes the entropy over semantic clusters to measure meaning-level uncertainty.

\noindent$\rhd$\textbf{Lexical Similarity}~\cite{lexicalsimilarity} measures the surface-level agreement among multiple sampled responses, where lower similarity indicates higher generation instability.

\ding{183}\textbf{Latent-based methods.} These methods exploit internal representations of LLMs to identify factuality-related signals beyond surface outputs. They are included to examine whether static hidden-state features are sufficient for detecting hallucinations in D-LLMs without explicitly modeling the denoising trajectory.

\noindent$\rhd$\textbf{EigenScore}~\cite{eigenscore} detects hallucinations by measuring the semantic consistency of internal representations, using the spectral structure of hidden-state covariance matrices.

\noindent$\rhd$\textbf{Contrast-Consistent Search (CCS)}~\cite{ccs} discovers truth-related directions in the activation space without direct supervision by enforcing consistency between contrastive statements.

\noindent$\rhd$\textbf{Truthfulness Separator Vector (TSV)}~\cite{tsv} learns a lightweight steering vector in the latent space to improve the separation between truthful and hallucinated responses.

\ding{184}\textbf{Trajectory-based methods.}
These methods are specifically designed for D-LLMs and leverage intermediate information exposed during iterative denoising. They provide the most direct comparison to \ours, since they also use process-level evidence rather than relying only on final outputs or static representations.

\noindent$\rhd$\textbf{TraceDet}~\cite{tracedet} is designed for D-LLMs and formulates the denoising process as a decoding trace. It selects hallucination-relevant sub-traces from intermediate denoising steps for final prediction.

\noindent$\rhd$\textbf{DynHD}~\cite{dynhd} constructs semantic-aware uncertainty evidence and learns reference denoising dynamics for factual responses. Hallucinations are detected by measuring the deviation between observed evidence trajectories and the learned reference trajectories.
\begin{table}[t]
\centering
\label{tab:hyperparameters}
\begin{tabular}{ll}
\toprule
\textbf{Hyperparameter} & \textbf{Value} \\
\midrule
Number of latent variables $K$ & 32, 48 \\
Hidden dimension $d$ & 64 \\
Fourier frequencies $F$ & 16 \\

\midrule
Cross-variable attention heads & 2 \\
Temporal attention heads & 8 \\
Dropout rate & 0.05 \\

\midrule
Optimizer & AdamW \\
Learning rate & $2\times10^{-4}$\\
Weight decay & 0.1 \\
Batch size & 8 \\
Maximum epochs & 100 \\
Early stopping patience & 20 \\
Trajectory preservation weight & 0.1 \\
Random seed & 42 \\
\bottomrule
\end{tabular}
\caption{Hyperparameter settings of DeMTS.}
\label{hyperparameter_set}
\end{table}

\begin{table*}[htbp]
\centering
\small
\renewcommand{\arraystretch}{1.2}
\begin{tabularx}{\textwidth}{l l X}
\toprule
\textbf{Dataset} & \textbf{Type} & \textbf{Question \& Response} \\
\midrule
\textit{HotpotQA}
& \textcolor{blue}{Factual}
& \textbf{Q:} Professor A. Selvanathan is a professor at a university
that is public or private? \\
& &
\textbf{R:} The university is \textcolor{blue}{public}. \\
\cmidrule{2-3}
&
\textcolor{red}{Hallucination}
&
\textbf{Q:} Which French aristocrat and military officer who fought in
the American Revolutionary War visited the historic home located at
Enfield, Halifax County, North Carolina, known as The Cellar? \\
& &
\textbf{R:} \textcolor{red}{James Hogun} visited The Cellar.
The reference answer is Marquis de Lafayette. \\
\midrule
\textit{TriviaQA}
&
\textcolor{blue}{Factual}
&
\textbf{Q:} What country has the largest Muslim population? \\
& &
\textbf{R:} The country with the largest Muslim population is
\textcolor{blue}{Indonesia}. \\
\cmidrule{2-3}
&
\textcolor{red}{Hallucination}
&
\textbf{Q:} Which supposedly non-lethal weapon was named by the inventor
after his childhood hero ``Tom Swift''? \\
& &
\textbf{R:} The weapon was called the
\textcolor{red}{Tom Swifty Gun}. \\
\midrule
\textit{CommonsenseQA}
&
\textcolor{blue}{Factual}
&
\textbf{Q:} The townhouse was a hard sell for the realtor; it was right
next to a high-rise what?

\textbf{Choices:}
(A) suburban development;
(B) apartment building;
(C) bus stop;
(D) Michigan;
(E) suburbs. \\
& &
\textbf{R:} \textcolor{blue}{B. apartment building}. \\
\cmidrule{2-3}
&
\textcolor{red}{Hallucination}
&
\textbf{Q:} There is a star at the center of what group of celestial
bodies?

\textbf{Choices:}
(A) Hollywood;
(B) skyline;
(C) outer space;
(D) constellation;
(E) solar system. \\
& &
\textbf{R:} \textcolor{red}{A. Hollywood}.
The correct answer is E. solar system. \\
\bottomrule
\end{tabularx}
\caption{Examples of factual and hallucinated responses from HotpotQA,
TriviaQA, and CommonsenseQA. Correct responses are highlighted in blue,
while hallucinated responses are highlighted in red.}
\label{table:hallucination_case_studies}
\end{table*}

\subsection{Implementation Details}\label{implementation}

We conduct experiments on TriviaQA, HotpotQA, and CommonsenseQA (show in Table~\ref{table:hallucination_case_studies}) using LLaDA-8B-Instruct and Dream-7B-Instruct as D-LLM backbones. For each dataset, we sample 1,600 training examples and split an additional 400 examples into 200 validation and 200 test instances. The random seed is fixed to 42. The number of denoising steps is set to 64 or 128. At each denoising step, we record token-wise predictive entropy and token IDs.

Our detector uses a hidden dimension of 64 and constructs (K=32,48) latent variables from the denoising trajectory. It is trained with Cross Entropy and the trajectory-preservation loss. We use AdamW with a learning rate of $2\times10^{-4}$, weight decay of 0.1, and batch size of 8. Training lasts for at most 100 epochs and early stopping with a patience of 20. The checkpoint with the highest validation AUROC is selected for testing. AUROC is used as the primary evaluation metric. All experiments are implemented in PyTorch 2.6.0 and conducted on 4 NVIDIA L40 GPUs. All experimental hyperparameter settings are shown in Table \ref{hyperparameter_set}.

\subsection{Sensitivity Analysis of Trajectory Preservation Losses}\label{losssensitivity} 
\begin{figure}[h]
\centering
\includegraphics[width=\linewidth]{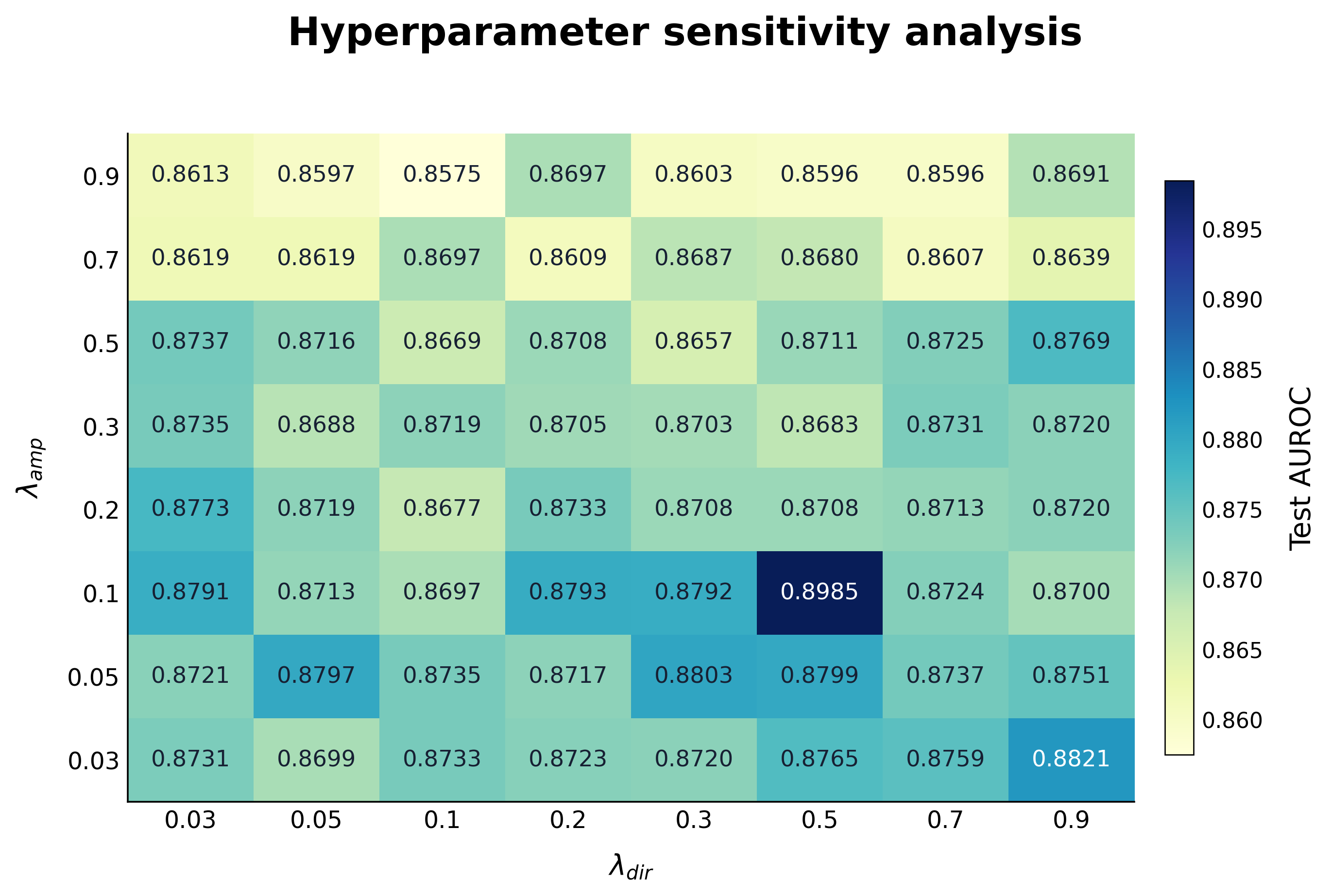}
\caption{Sensitivity analysis of the amplitude consistency loss $\mathcal{L}_{amp}$ and direction consistency loss $\mathcal{L}_{dir}$.} 
\label{figure:loss_sensitivity}
\end{figure}

We further analyze the sensitivity of the two trajectory-preservation losses, i.e., the amplitude consistency loss $\mathcal{L}_{amp}$ and the direction consistency loss $\mathcal{L}_{dir}$. As shown in Fig ~\ref{figure:loss_sensitivity}, the detector maintains stable performance across a broad range of loss weights, indicating that the proposed trajectory-preservation objective is not overly sensitive to specific hyperparameter choices. When either loss is removed or assigned a very small weight, the performance decreases, suggesting that preserving only the classification objective is insufficient to retain fine-grained denoising dynamics during token-to-variable assignment. Meanwhile, excessively large weights also lead to performance degradation, since overly strong trajectory constraints may suppress discriminative feature learning for hallucination detection. Overall, the heatmaps show that $\mathcal{L}_{amp}$ and $\mathcal{L}_{dir}$ play complementary roles: $\mathcal{L}_{amp}$ helps preserve the magnitude of local uncertainty variations, while $\mathcal{L}_{dir}$ maintains the temporal direction of denoising evolution. Their joint use provides a stable and effective regularization scheme for constructing reliable variable-level trajectories.

\subsection{Complete Ablation Results}\label{appendix:complete_ablation}

To provide a more comprehensive analysis of each component in \ours, we report the complete ablation results on both D-LLMs backbones in Table~\ref{tab:full_ablation}. The results exhibit consistent trends across different models, datasets, and generation-length settings, further verifying the robustness and generality of our design.

The complete ablation results further confirm the effectiveness of each component in \ours. Replacing schedule-normalized trajectories with raw uncertainty signals consistently reduces performance, showing the importance of suppressing schedule-induced global trends. Directly using token positions as variables also leads to clear degradation, verifying the necessity of constructing semantically stable latent variables before time-series modeling. Removing either $\mathcal{L}_{amp}$ or $\mathcal{L}_{dir}$ also weakens performance, indicating that both amplitude consistency and direction consistency contribute to preserving fine-grained denoising dynamics during token-to-variable assignment. Moreover, removing inter-variable interaction or temporal evolution modeling consistently degrades performance, demonstrating that \ours benefits from jointly preserving local denoising dynamics and modeling multivariate temporal dependencies.

Overall, the complete ablation results are consistent with the observations in the main paper. They demonstrate that the performance gain of \ours does not come from a single isolated component, but from the integration of schedule normalization, stable latent-variable construction, trajectory-preserving regularization, and dynamic multivariate temporal modeling. 

\section{Prompt Templates}
\subsection{D-LLM Response Generation}

We design task-specific generation prompts according to the answer format and evidence setting of each benchmark. TriviaQA requires concise factual answers, HotpotQA requires responses grounded in the provided context, and CommonsenseQA requires selecting a single option without additional explanations. Across all datasets, the final answer is enclosed within \texttt{<answer></answer>} tags to enable consistent answer extraction and subsequent hallucination annotation. The complete prompt templates are provided below.
\begin{generationprompt}{TriviaQA Generation Prompt}
Answer the question concisely. Question: {question}

And please put your final answer in <answer> </answer>
\end{generationprompt}

\begin{generationprompt}{HotpotQA Generation Prompt}
You are given the following context{context}

Question: {question}
Answer the question based on the context only.
Please put your final answer in <answer> </answer>
\end{generationprompt}

\begin{generationprompt}{CommonsenseQA Generation Prompt}
Question: {question}

Options:
A. {option_a}
B. {option_b}
C. {option_c}
D. {option_d}
E. {option_e}

Instruction:
- Select exactly ONE correct option (A, B, C, D, E).
- DO NOT generate explanations.
- Output format MUST be: <answer>X</answer>,
  where X is one of {A, B, C, D, E}.
- Any other output will be considered invalid.

Your output:
\end{generationprompt}

\subsection{Automatic Hallucination Annotation}

For TriviaQA and HotpotQA, we use Qwen3-8B to automatically determine
whether the generated response is hallucinated. 
For TriviaQA and HotpotQA, the judgment \texttt{yes} is mapped to
label \(1\), indicating a hallucinated response, whereas
\texttt{no} is mapped to label \(0\), indicating a
non-hallucinated response. CommonsenseQA does not use the Qwen3-8B
judge; its labels are obtained through rule-based matching of the
predicted option or answer text.
\begin{judgeprompt}{Hallucination Evaluation Instructions}
You are a helpful assistant.
Your task is to determine whether the output contains hallucination.
Follow these guidelines strictly:

Fluency Check:
If the output is not fluent natural language
(e.g., it contains garbled or unreadable text),
it should be considered hallucinated.

Relevance Check:
If the output contains many correct facts but does not directly
answer the question, it should be considered hallucinated.

Support Check:
If the output cannot be inferred from any of the reference
answers, or contains information inconsistent with the reference
answers, it should be considered hallucinated.

Exact Match Rule:
If the output is supported by any one of the reference correct
answers, it should be considered not hallucinated.

Semantic Match Rule:
If the output is not directly supported by any reference answer,
but is semantically similar, i.e., it expresses the same meaning,
it should be considered not hallucinated.

Unknown Answer Rule:
If the reference answers include phrases such as
"This question cannot be answered", then an output such as
"I don't know" or "Cannot answer this question" should be
considered not hallucinated.

I understand. Please provide the question and the bot's answer.

Question: {question}

The correct answer examples are as follows:
{reference_answer_1}
{reference_answer_2}
...
The bot replied as follows:
{model_answer}

Now please judge whether the bot's answer is hallucinated or not.
If it is hallucinated, please answer "yes"; otherwise, answer "no".
Do not show your reasoning, and put your answer in
<answer> </answer>.
\end{judgeprompt}

\end{document}